%% file: main.tex
\documentclass[manuscripts,creen]{acmart}
\AtBeginDocument{%
  \providecommand\BibTeX{{%
    \normalfont B\kern-0.5em{\scshape i\kern-0.25em b}\kern-0.8em\TeX}}}

\setcopyright{none}
\copyrightyear{2023}
\acmYear{2023}
\acmDOI{}

\acmConference[FAccT '23]{ACM Conference on Fairness, Accountability, and Transparency}{June 12--15, 2023}{Chicago, IL}

\usepackage{caption}
\usepackage{subcaption}
\usepackage{tabulary}
\usepackage{amsthm}
\usepackage{amsmath}
\usepackage{multirow}

\usepackage[textsize=small]{todonotes}

\newtheorem*{remark}{Remark}

\begin{document}

\title{An Operational Perspective to Fairness Interventions: Where and How to Intervene}

\author{Brian Hsu}
\email{bhsu@linkedin.com}
\affiliation{%
  \institution{LinkedIn}
  \city{Sunnyvale}
  \state{CA}
  \country{USA}
}

\author{Xiaotong Chen}
\email{elachen@linkedin.com}
\affiliation{%
  \institution{LinkedIn}
  \city{Sunnyvale}
  \state{CA}
  \country{USA}
}

\author{Ying Han}
\email{yihan@linkedin.com}
\affiliation{%
  \institution{LinkedIn}
  \city{Sunnyvale}
  \state{CA}
  \country{USA}
}

\author{Hongseok Namkoong}
\email{namkoong@gsb.columbia.edu}
\affiliation{%
  \institution{Columbia University and LinkedIn}
  \city{New York}
  \state{NY}
  \country{USA}
}

\author{Kinjal Basu}
\email{kbasu@linkedin.com}
\affiliation{%
  \institution{LinkedIn}
  \city{Sunnyvale}
  \state{CA}
  \country{USA}
}


\begin{abstract}
    As AI-based decision systems proliferate, their successful operationalization requires balancing multiple desiderata: predictive performance, disparity across groups, safeguarding sensitive group attributes (e.g., race), and engineering cost. We present a holistic framework for evaluating and contextualizing fairness interventions with respect to the above desiderata. The two key points of practical consideration are \emph{where} (pre-, in-, post-processing) and \emph{how} (in what way the sensitive group data is used) the intervention is introduced. We demonstrate our framework with a case study on predictive parity. In it, we first propose a novel method for achieving predictive parity fairness without using group data at inference time via distibutionally robust optimization. Then, we showcase the effectiveness of these methods in a benchmarking study of close to 400 variations across two major model types (XGBoost vs. Neural Net), ten datasets, and over twenty unique methodologies. Methodological insights derived from our empirical study inform the practical design of ML workflow with fairness as a central concern. We find predictive parity is difficult to achieve without using group data, and despite requiring group data during model training (but not inference), distributionally robust methods we develop provide significant Pareto improvement. Moreover, a plain XGBoost model often Pareto-dominates neural networks with fairness interventions, highlighting the importance of model inductive bias. 
\end{abstract}





\maketitle
\input{Sections/Intro}
\input{Sections/Framework}
\input{Sections/Background}
\input{Sections/CaseStudy}
\input{Sections/Conclusion}
\clearpage
\appendix

\bibliographystyle{ACM-Reference-Format}
\bibliography{fairness}
\input{Sections/Appendix}
\end{document}

%% file: Sections/Intro.tex
\section{Intro}
\subsection{Holistically Approaching Algorithmic Fairness}
As machine learning (ML) models increasingly play a role in day-to-day life, algorithmic fairness has been steadily gaining attention in both public and academic eyes. Fortunately, there has been no shortage of research into measuring and achieving different definitions of algorithmic fairness, with a recent set of comprehensive surveys (\cite{ChenTestingSurvey, HortMitigationSurvey}) describing and categorizing over 300 fairness intervention\footnote{The authors of the survey call these methods "bias mitigation" methods and semantics vary across institutions. For this work, we will refer to all algorithms with the end goal of inducing some measure of fairness to be "fairness interventions" or just "interventions."} algorithms and 100 testing algorithms. However, alongside the spectrum of such algorithms comes the practical problem of selecting the "right" intervention strategy in real-world applications. Several recent works have recognized that this is a critical choice that stretches beyond sole consideration of the canonical fairness effectiveness versus model performance trade-off. We contribute to this literature by proposing a selection framework that emphasizes two additional key dimensions for evaluating fairness interventions. The first dimension is \textit{how} the algorithm is applied, specifically in terms of its sensitive demographic data requirements\footnote{Examples include ethnicity, gender identity, etc. We hereafter refer to these general attributes as "group data."}, and the second dimension is with respect to \textit{where} the intervention is applied between pre-/in-/post-processing, which is closely linked to engineering effort in real settings. We identified these aspects as critical to operationalization as they have major implications on the public and legal perception of the fairness intervention as well as the engineering scalability of the solution. 

To concretely motivate the problem, consider the commonly-used categorization of fairness intervention algorithms into three categories - pre-processing, in-processing, and post-processing, which respectively correspond to adjustments of the data, model, and model outputs. Post-processing solutions are effective in that they directly tackle the fairness problem at the decision level and are considered scalable from an engineering perspective as they can be readily detached from and/or applied onto models with little to no fine-tuning. However interventions at this level typically use the demographic information (e.g., gender, race, religion) as part of the output adjustment (see \cite{Hardtetal_NIPS2016, pleiss2017fairness, NandyEOdds, Hsu2022} for examples). On the other hand, interventions at the in-processing stage may use the demographic information in regularization fashion (e.g, \cite{Celis2019, cotterNonconvexConstrained, Rezaei_2020}). While these approaches may be harder to implement and tune across many models (and therefore arguably lack engineering scalability), the critical difference from an algorithmic perspective is that \emph{these in-processing methods do not require using the demographic group data at inference time while many post-processing schemes do require the data}. 

\begin{center}
    \begin{figure}[h] 
    \caption{Our proposed framework for operationalizing fairness. }
    \label{fig:Framework}
    \includegraphics[width=\textwidth]{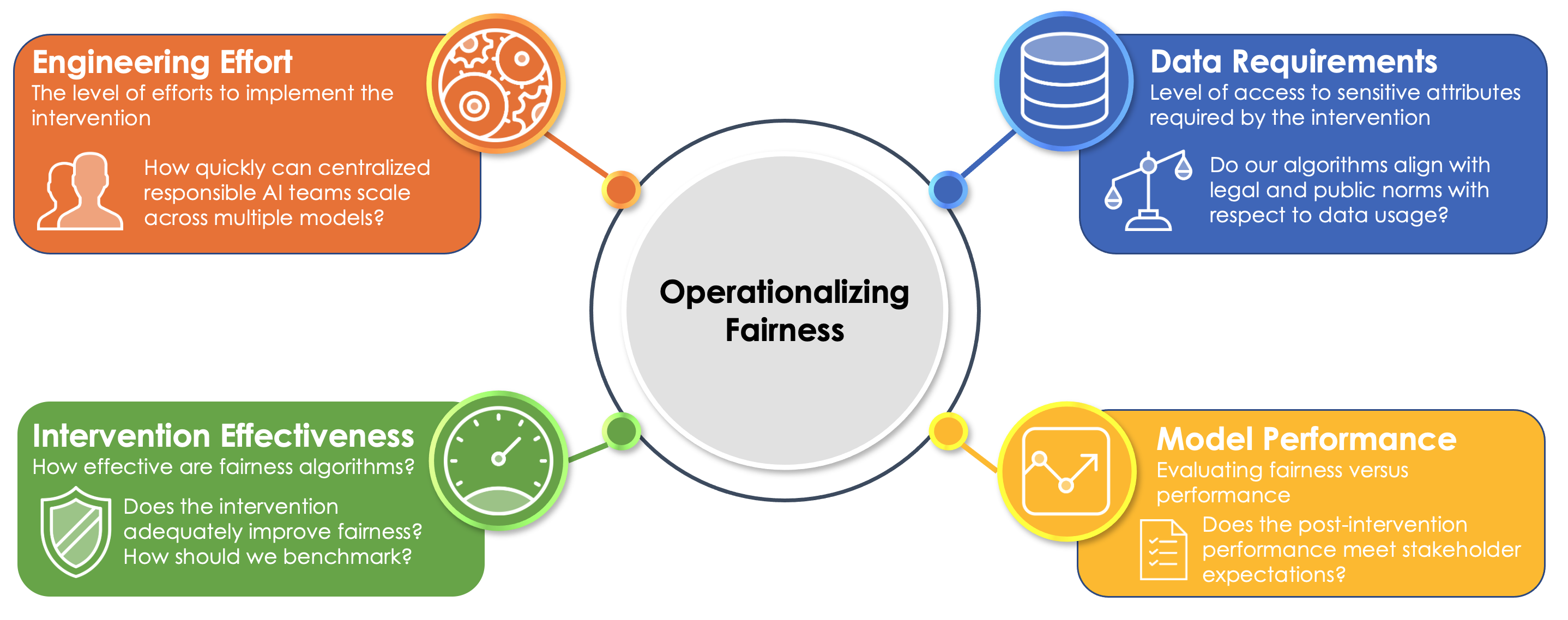}
    \end{figure}
\end{center}

The collection and use of group data is a complex topic that involves legal, ethical, and organizational considerations. Hence, in many practical settings, group data is often unavailable or costly to collect. Furthermore, organizations that do have such data often have stringent access controls due to its sensitivity. In typical tech stacks, these factors translate into operational constraints on which interventions can be  deployed, and the level of effort required to implement them. Hence, on top of the traditional trade-offs between model performance and fairness, we introduce two important axes to consider in industrial applications---data availability and engineering effort. We capture these ideas in Figure \ref{fig:Framework}.

\subsection{Previous Work and Contributions}
Several recent works have drawn attention to the complexities of fairness in ML beyond the algorithms. These works bring in interdisciplinary perspectives on fairness, including arguments from the legal \cite{Bringas_Colmenarejo_2022, bent2019algorithmic, xiang2020reconciling}, economic \cite{rambachan2020economic, AndrusDemographic}, moral \cite{Loi2022IsCA, Bringas_Colmenarejo_2022}, and industrial \cite{pfohl2021empirical, BogenAwareness, bakalarFairnessOnTheGround, AndrusSpitzer} perspectives. These works highlight the need to concretely evaluate how group data is used for fairness purposes and to what extent interventions work without such data. . The essay by Andrus and Villeneuve \cite{AndrusDemographic} is of particular pertinence to our work, as it calls for practitioners to think twice before blindly pursuing data gathering at all costs due to the hazards ranging from organizational risks to individual hazards to the organization's customers/members.  
Many of these works instill much-needed grounding perspectives at the higher system/organization level\footnote{By system level, we are referring to broad directives such as how group data can be collected from users, differences between human and algorithmic bias, who deserves the benefit from algorithmic fairness, etc. This is as opposed to granular decisions at the model or product level such as selecting a specific point of intervention or even a specific fairness intervention algorithm itself.}. However, there is now a need to bridge these system level directives with the growing arsenal of fairness intervention methods. Some examples of such investigations are \citet{vzliobaite2016using}, which theoretically argues for the use of group data during model training and demonstrates it on salary data. Another example is \citet{grgic2018beyond}, which introduces and addresses the problem of process fairness and demonstrates its on data sourced from Amazon's Mechanical Turk. The latter is particularly relevant for our work, as it analyzes the human perceived fairness of using individual features and solves the problem of finding a fair and good enough subset of features. Likewise, we emphasize in our work that fairness choices can and should be viewed as a spectrum rather than Boolean switches. Lastly, another remarkable work is \citet{Mendez2022-MENEIO}, which looks at the fairness differences of intervening at the in- or post- processing stage, which at face value appears to be an engineering decision. The authors highlight that even when pursuing the same fairness criterion, the explicit reliance on group data in post-processing methods, compared to the implicit reliance of in-processing during model training, can lead to significantly different predictions for the same individual. We emphasize the importance of this choice even further and argue that the more pressing practical issues are that of the data requirements and engineering efforts behind these different points of intervention.

Following in these footsteps, our work makes two main contributions that tie the system level fairness discussions down to quantitatively choosing between fairness interventions. First, we propose a framework that highlights four key pillars for selecting a fairness intervention solution that holistically considers model performance and algorithmic effectiveness alongside two operationally critical axes - data requirements (how do we intervene) and engineering effort (where do we intervene).  Our qualitative insights and practical heuristics allow navigating the trade-offs between these pillars in industrial settings.

To quantitatively showcase these trade-offs, we implement and evaluate a range of fairness intervention algorithms (see Figure \ref{fig:Aggregate ACS Frontier}). We demonstrate our framework with a case study on predictive parity. In it, we first propose a novel method for achieving predictive parity fairness without using group data at inference time via distibutionally robust optimization. Then, we showcase our proposed framework in a benchmarking study of close to 400 variations across two major model types (XGBoost vs. Neural Net), ten datasets, and over twenty unique methodologies. Methodological insights derived from our empirical study inform the practical design of ML workflow with fairness as a central concern. We find predictive parity is difficult to achieve without using group data, and despite requiring group data during model training (but not inference), distributionally robust methods provide significant Pareto improvement. Moreover, a plain XGBoost model often Pareto-dominates neural networks with fairness interventions, highlighting the importance of model inductive bias in fairness.

%% file: Sections/Framework.tex
\section{Holistic Fairness Intervention Selection Framework} \label{Section: Framework}

We now propose our framework for operationalizing fairness interventions in practice. The key objective of the framework is to tie the system-level objectives and constraints (e.g., degree of access to group data, engineering effort and budget) to the selection of the right fairness intervention algorithm. This is achieved by emphasizing two major aspects of fairness intervention---the \emph{how} and the \emph{where}. They dictate the class of the candidate algorithms and encapsulate both quantitative (data size/availability and intervention model complexity) and qualitative (engineering effort and how transparent  the intervention is to stakeholders) considerations in ML development. We will present our framework in four pillars, with the first two focusing on these operationally-focused aspects and the latter focusing on the canonical considerations of fairness versus average predictive performance trade-offs. In describing our proposed evaluation guidelines, we focus our guidance for usage in large-scale online platforms; we expect our development will often have a bigger bearing on non-technology companies with less sophisticated engineering infrastructures. Note that while the choice of which fairness definition to use is a major aspect of operationalization, we consider it to be out of scope as it is highly context dependent. 

\paragraph{\textbf{Group Data Requirements (How)}}
\begin{itemize}
    \item \textit{What is it}: We propose that this axis ranges from group data availability in (in order of most to least preferable) - Nowhere, validation data, training \& validation data, training \& validation \& inference data.
    \item \textit{Why it matters}: The collection of group data itself can be a point of contention. As \citet{AndrusSpitzer, bent2019algorithmic, xiang2020reconciling} point out, the legality of collecting and using group data in models is ambiguous from a discrimination perspective, even if only for auditing purposes. On top of the legal risks, \citet{AndrusDemographic} argue that group data collection, even for honest purposes of inducing fairness, can result in harmful social effects. On the other hand, \citet{BogenAwareness, vzliobaite2016using} present cases where group data is necessary and beneficial for algorithms to be fair. Overall, this is to say that there does not seem to be a one-size-fits-all ruling for using group data. 
    
    This motivates us to move beyond the dichotomous view of viewing group data as fully available or not. We consider the graduated scale of group data, which enables more flexible and complex fairness intervention algorithms to be used. As a concrete example, obtaining a small set (validation set) of group data from volunteers may enable practitioners to tune models for fairness. If the data is large enough, it may even permit training a model with group data that does not require access at inference time. On the extreme end, having full access to group data at training and inference would enable using a variety of post-processors. As \citet{AndrusSpitzer,grgic2018beyond} discuss, these different regimes can also have a graduated on public perception. The matter of perception is further nuanced by the fact that access to group data may be restricted select practitioners within an organization \cite{nandy_2021}. We therefore recommend practitioners to quantitatively study how far they can get under the different data availability regimes to justify the use of group data. We specifically demonstrate this process in our case study of predictive parity in Section \ref{section: Case Study}. 

\end{itemize}

\paragraph{\textbf{Place of Intervention and Engineering Effort (Where)}}
\begin{itemize}
    \item \textit{What is it}: We propose that the current categorization of pre-/in-/post- processing is a clear stratification of where the intervention occurs and that each carries its pros and cons in practice with respect to engineering effort and scalability.
    \item \textit{Why it matters}: First and foremost, the engineering effort can vary widely between the three levels.  Post-processing interventions typically allow lightweight iteration, and AB testing between existing (base) models versus the base model with the post-processor attached, as the model can effectively be treated as an opaque box. For example, a post-processing tool for calibration (e.g. isotonic regression) has relatively few parameters to train and only requires only understanding of the model outputs and labels.
    
    On the other hand, in-processing mitigation tends to require a deeper understanding of the base model architecture (see the recent methodological survey \cite{HortMitigationSurvey} for examples). For instance, some regularizers that work for one model may not work for another due to dependence on assumptions such as convexity or differentiability. Additionally, more engineering effort may be required to push customized changes to training schemes, loss functions, or tune the associated hyperparameters, which lengthens time to experimentation. 
    
    The effort is further complicated when moving to pre-processing schemes, which tend to require a very fine-grained understanding of model details and features. In this regime, there are very fundamentally difficult factors to consider here as opposed to in-/post- processing; for instance, consider the distinction between numeric, categorical, and embedding variables. While understanding the bias of features is undeniably important \cite{grgic2018beyond}, it is effort-intensive, requires ad-hoc qualitative judgment, and is difficult to scale across organizations. 

    The aforementioned engineering difficulties compound as many large-scale software companies opt for a centralized entity to handle responsible/fair/ethical AI affairs (hereafter, "responsible AI teams") \cite{kramer_2021, logan_2022, kloumann_tannen_2021, croak_gennai_2023}. A centralized entity has the significant benefit of mandating consistent responsible AI principles and developing cohesive tooling. However, its main drawback is that highly customized solutions become difficult to implement and scale across the company.
    
    For transparency purposes, we find from the fairness intervention survey \cite{HortMitigationSurvey} that post-processing frameworks tend to enable finer control and better explainability than in-processing. For instance, whereas it is difficult to understand how strong in-processing regularizers need to be to achieve a certain effect, post-processing algorithms tend to resemble constrained optimization problems that adjust the base models' scores (or nearly equivalently, classification thresholds) with interpretable parameters (\citet{HortMitigationSurvey} mentions at least 20 of such examples). Having this transparency would largely aid in explaining the intervention to various stakeholders (e.g., legal and product teams, as well as users). 
    
    To summarize, the qualitative consideration of where the intervention occurs poses significant practical consequences. Both researchers and practitioners must think ahead on the question of \textit{where} for the sake of engineering effort/scalability and transparency, tailoring to the idiosyncracies of each organization.
\end{itemize}  

\paragraph{\textbf{Intervention Effectiveness}}
\begin{itemize}
    \item \textit{What is it}: Intervention effectiveness refers to how well a chosen methodology achieves its fairness goal. For instance, if pursing an equalized odds type of fairness, one may look at the largest largest differences in TPR/FPRs across the groups. An important distinction here is also for whom the intervention should work for (e.g. group versus individual fairness). In our case study of predictive parity fairness, we will measure the worst group calibration error. 
    \item \textit{Why it matters}: This is the typical end-goal of the fairness practitioner. In measuring effectiveness, we encourage practitioners and researchers to do so with respect to the above two points. A post-processing algorithm that requires group data at inference may yield better fairness effects than an in-processing algorithm. However, whether the improvement is worth the extra data is an empirical problem that we believe requires benchmarking to answer. \citet{HortMitigationSurvey} indicates that most benchmarks are done with respect to the same category of interventions (in-/pre-/post-). However, some are done across categories and it is not clear if such benchmarks are consistent in the amount of group data being used. We therefore encourage that given the practical importance of data requirements, benchmarks in academic and industrial settings be done with respect to both intervention category and the data requirements. 
\end{itemize}  

\paragraph{\textbf{Model Performance}}
\begin{itemize}
    \item \textit{What is it}: Model performance refers to the difference in model performance compared to the baseline when applying a fairness intervention. This may typically be measured by a performance metric such as accuracy, but other times may involve downstream metrics in products such as click-through-rate. 
    \item \textit{Why it matters}: Demonstrating an adequate level of model performance post-intervention is typically a requirement to get buy-in from product teams and other stakeholders. Although some recent works have shown empirically that the fairness versus performance trade-off is not as steep as one may assume \cite{Rodolfa_2021}, it still remains a valid concern. It appears that this is a well-known factor and we encourage practioners to continue quantitatively benchmarking fairness interventions on the basis of model performance.  
\end{itemize}  

This framework overall aims to capture the major considerations in operationalizing fairness that are quantifiable and enable benchmarking to some extent, as we believe that this helps practitioners decide how to make trade-offs between the pillars. We remark that while these are the classical categorizations in ML pipelines, there are still applications that use group data in ways that fall outside of these categories (for instance at the data collection step \cite{caiAdaptiveSampling, li2022more}, we also propose one such method in the next section at the feature selection step). These methods should be considered, but our work focuses on bringing more structure and order to the majority of fairness intervention work in the highlighted categories \cite{HortMitigationSurvey}. With this four pillar framework in mind, we pivot our work to demonstrating this on a case study of predictive parity fairness.

%% file: Sections/Background.tex
\section{A Primer on Predictive Parity Fairness}
For our case study, we will focus on a particular group fairness definition known as predictive parity (hereafter abbreviated as PP), also known as sufficiency. In this section, we first provide background on PP and motivate its importance. Then, we describe how and why we measure PP through the worst-group calibration error of a model. Lastly, we provide an overview of methods for achieving PP that recaps well-known methods and also introduces new methods inspired by robust optimization principles. 

\subsection{Honing in on Predictive Parity}
PP as a definition was popularized in 2017 in the context of the COMPAS Recidivism study \cite{dieterich2016compas}\cite{Chouldechova17}. In words, this condition states that a model predicted score should have the same interpretation of outcome likelihood regardless of who the prediction is being made for. Formally, it states that for a classifier that produces a risk score $s\in [0,1]$ to predict an observation's outcome propensity $Y\in\{0,1\}$ where the observation's group membership $g\in \mathcal{G}=\{1,\dots,G\}$, we must have the following:
\begin{equation}
\label{eq:pp} E[Y|S=s, G=i] = E[Y|S=s, G=j] \quad \forall i,j \in \{\mathcal{G}| i\neq j \} 
\end{equation}

PP is closely related to notions of equal calibration \cite{Chouldechova17,barocas-hardt-narayanan}, which is a stronger notion that states that the score produced by a model is representative of the empirical likelihood of the event it is predicting, regardless of who the prediction is for. Formally, this is the condition that: 
\begin{equation}
\label{eq:equalCal} E[Y|S=s, G=g]=s \quad \forall g\in \mathcal{G}
\end{equation}
We will return to this connection in the next subsection but for now reference equal calibration to motivate the importance of achieving PP. There are several reasons why PP is a desirable fairness property in a variety of domains. First, it is a notion of fairness that is easily justifiable as it does not explicitly harm or benefit one group, but rather tries to ensure that scores are reflective of empirical risk. Second, as ML models are often used in junction with human decision-makers, equal calibration ensures that model outputs are interpretable and comparable. Lastly, calibration \eqref{eq:equalCal} is a desirable property when scores are used in safety-critical applications such as healthcare \cite{pfohl2021empirical} or when used in downstream purposes such as candidate ranking \cite{nandy_2021}.

\subsection{Predictive Parity and Calibration} \label{subsection: PP and Cal}
As elaborated in \cite{barocas-hardt-narayanan}, there is a clear linkage between PP and calibration. Namely, a model that satisfies PP can be made calibrated through a simple output mapping transformation, and on the flip side, a calibrated model satisfies PP. This linkage thus provides a straightforward method to achieving PP by simply calibrating the model. This approach is favored as it also brings the aforementioned benefits of calibration. Hence, if we accept that calibration is the most intuitive strategy to achieve PP, then evaluating for PP fairness boils down to ensuring that each group is equally well calibrated. In other words, we aim to ensure that the fairness intervention for predictive parity through calibration worked equally well for all groups. 

Thus, we define PP in the multiple group setting for our study by looking at the measuring the calibration error (CE) of the worst-calibrated group on the testing data $\mathcal{D}$, which is comprised of model scores $s$, binary labels $y$, and group labels $g$.
\begin{equation}
    Worst \: group \: CE = \underset{g\in\mathcal{G}}{max}\ CE_{(s,y,g)\sim \mathcal{D}}(f) \label{eq: wgce}
\end{equation}

This is as opposed to the average or overall CE: $CE_{(s,y,g)\sim \mathcal{D}}(f)$, which may overlook the (in)effectiveness of the fairness intervention to any particular group. As for which CE we will use, we opt for the Expected Cumulative Calibration Error (ECCE) proposed in  \citet{arrieta2022metrics}. This metric was selected over the commonly-used expected calibration error (ECE) \cite{guoCalibration} after we surveyed recent developments in CE metrics and found several works pointing out that ECE had statistical and empirical flaws. We recap these arguments in detail in Appendix \ref{App: CE Metrics} and summarize our case study for two other CE metrics we considered as well (the mean sweep calibration error (MSCE) of \cite{roelofs2022mitigating} and the kernel calibration error (KCE) of \cite{BlasiokCalibration}) in Appendix \ref{App: Results for MSCE}, \ref{App: Results for KCE}. With our goal established as investigating methods for minimizing worst-group ECCE for predictive parity fairness, we now turn to our case study. 

\subsection{Methods for Achieving Predictive Parity}

Existing works for obtaining PP include \citet{pleiss2017fairness} which attempts to maintain calibration of a model alongside equalized odds, \citet{pmlr-v80-hebert-johnson18a} which presents a individual-fairness notion of calibration called multicalibration, \citet{Hsu2022} which uses group membership in a constrained-optimization style post-processor for PP but does not calibrate, \citet{LeeSufficiency} which regularizes for PP using an information-theoretic approach, and \citet{zeng2022sufficiency} which applies to binary $0/1$ classifiers.

In our review of algorithms and recent surveys \cite{HortMitigationSurvey}, methods for achieving PP does not appear to receive as much focus as other group fairness definitions such as demographic parity and equalized odds \cite{Hardtetal_NIPS2016} as also observed in \cite{LeeSufficiency} and \cite{zeng2022sufficiency}. The relative lack of attention may be due to "obviousness" of the solution, being to simply calibrate the model separately on each group or to use the group attribute. Moreover, there is theoretical evidence showing that PP essentially comes for free in unconstrained learning situations when the model features have enough predictive power to infer group membership \cite{LiuImplicitFairnessCriterion}. The relative ease of achieving PP when group membership is available as a feature has perhaps led to less attention on fairness intervention strategies. However, we believe that despite these arguments, further investigation into achieving PP is warranted. Firstly, as a central theme to this work, group data is not always available as a model feature due to optical or legal reasons. Moreover, even if it is available, there are reasons to avoid its usage. Secondly, the conditions mentioned in \cite{LiuImplicitFairnessCriterion} may not always be met as many commonly used models are \textit{not} trained with unconstrained loss minimization (e.g. regularized training) nor are features always informative of group membership. Hence, our work also aims to achieve PP without full group data access. 
\vspace{-0.5cm}
\begin{center}
    \begin{figure}[h] 
    \caption{Methods benchmarked in our case study analyzed by our proposed four pillars}
    \label{fig:POI}
    \includegraphics[width=\textwidth]{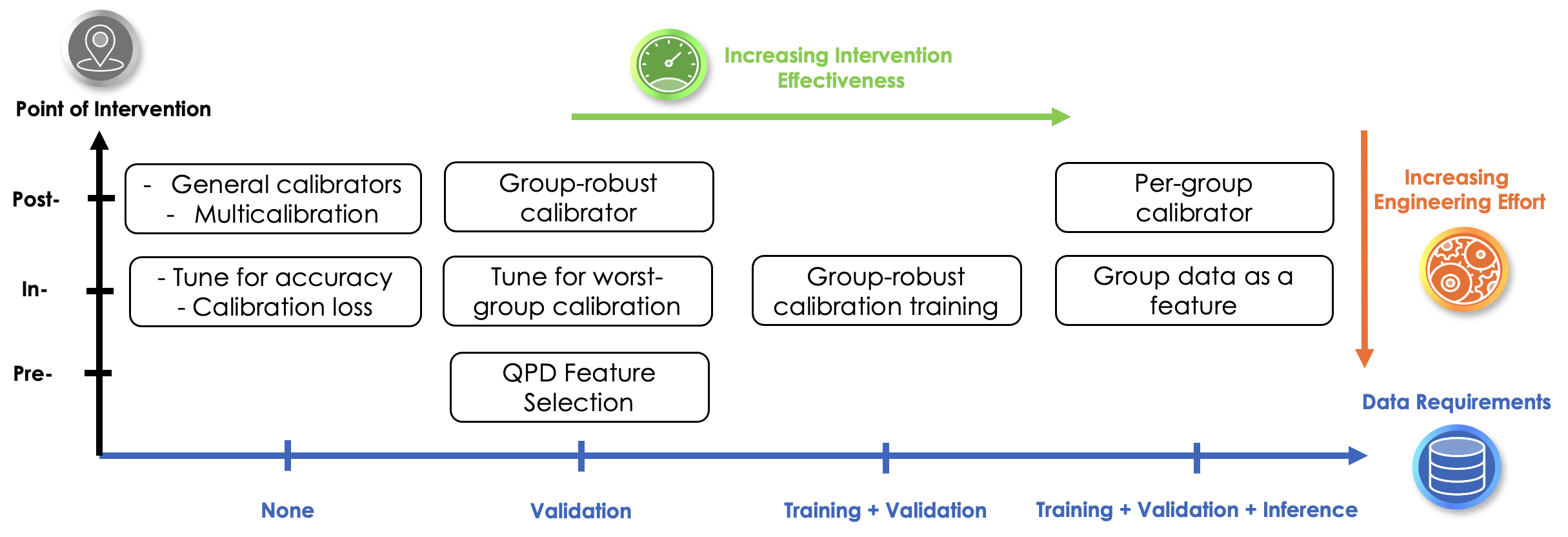}
    \end{figure}
\end{center}
\vspace{-0.5cm}
Towards this goal, our benchmarks cover a multitude of established solutions for achieving predictive parity (e.g., using the group attribute, model calibration, using a calibration loss, tuning for calibration). On top of this, we also devise a new in-processing method for predictive parity by relying on distributionally robust optimization. While DRO-methods have primarily focused on accuracy-type metrics \cite{GardnerXGB, sagawa2019distributionally, zhai2021doro, pmlr-v80-hashimoto18a}, we show that it can also be effectively adapted for other fairness definitions. As we will illustrate in Section \ref{section: Case Study}, this method consistently beats the baseline and non-robust counterpart. For all benchmarked methods, we depict them from the lens of our four-pillar framework in \ref{fig:POI}. Our methods focus on achieving PP in two models that are commonly used in industry for their consistently strong performance on tabular data - Extreme Gradient Boosting (XGB) \cite{Chen:2016:XST:2939672.2939785} and feed-forward neural networks (NN). Due to the large suite of methods we benchmark, we will provide a brief description of each methodology below, in decreasing order of data requirements and provide thorough descriptions with implementation details in the appendix. The ordering is meant to reflect the increasing "acceptableness" of the mitigation approach, as avoiding the use of demographic/protected attribute data is generally seen as more palatable according to the multiple desiderata discussed above. Another theme of these methods is the contrast of addressing average-case calibration error by avoiding use of group data versus addressing worst-group calibration error at the expense of requiring some extent of group data.

\begin{itemize}
    \item \textbf{Train + Validation + Inference}: In this regime, the protected attribute data is available for use in all steps of the model pipeline \footnote{If a methodology only requires group data at inference and validation (e.g. per-group calibrator, we consider the requirement of data at inference to be a stronger requirement than training hence we place it in this category.}.
    \begin{itemize}
        \item \textbf{Group as a feature} (In): We encode the group attribute as indicator variables (one for each group) and use them in the model at all steps. 
        \item \textbf{Per-group calibrator} (Post): After a model is tuned for accuracy, we use it to predict probabilities on the validation set and use the scores and labels to fit a post-hoc calibrator (isotonic regression \cite{MizilPlattIsotonic}) separately for each group. At inference, we use the group attribute of each observation to select the respective model for calibrating the scores. This explicitly aims to optimize for worst-group calibration error \eqref{eq: wgce}. 
    \end{itemize}
    \item \textbf{Train + Validation}: In this regime, the protected attribute is available in the training and tuning phase, but not accessible for inference. 
    \begin{itemize}
        \item \textbf{Group-robust calibration training} (In): This is a new method that we propose as an in-processing approach to achieving PP fairness. In it, we set the objective of the gradient boosting and neural network models to simultaneously optimize the base loss (cross entropy) and the worst-group calibration loss $CE(f)$: 
        \begin{equation}
            \mathcal{L}(x,y,g)=\sum_{i} y_{i}log(f(x_{i}))+(1-y_{i})log(1-f(x_{i}))+\gamma( \underset{g\in\mathcal{G}}{max}E_{(x,y)\sim\hat{P_{g}}}[CE(f)])
            \label{eq: droCalLoss}
        \end{equation} 
        While the first term is standard, the second term is achieved via distributionally robust optimization. The intuition is to have the model focus more on the group that has the worst calibration. For neural networks, the DRO loss \cite{sagawa2019distributionally} is applied on top of the Maximum Mean Calibration Error (MMCE) loss from \citet{kumar2018trainable}. For XGB, the DRO loss\footnote{We use DRO here loosely, as the exact mirror descent algorithm proposed in \citet{sagawa2019distributionally} can only be heuristically replicated in the XGB framework.} is applied on top of Brier score. We provide further details and commentary in Appendix \ref{App: Methodology list} for why we have chosen the respective loss functions. 
    \end{itemize}
    \item \textbf{Validation}: In this regime, the protected attribute is available for tuning purposes or training models with very few hyperparameters (such as calibrators that train on one variable). 
    \begin{itemize}
        \item \textbf{Tune for worst-group calibration} (In): For each set of hyperparameters, we evaluate the resulting CE on the validation data and pick the set of parameters that yields the smallest worst-group CE.
        \item \textbf{Group-robust Calibrator} (Post): We fit a boosting model except instead of using the original data features, the only feature is the predicted score from the base (uncalibrated) model. The model is trained with the mixture group-robust loss function. 
    \end{itemize}
    \item \textbf{None}: In this regime, we do not have the protected attribute at any stage.
        \begin{itemize}
            \item \textbf{Calibration loss} (In): The default choice for binary prediction loss functions is cross entropy/logistic loss. However, there are other loss functions that fare better for calibration purposes that have been explored in NNs. These include temperature scaling \cite{guoCalibration}, focal loss \cite{mukhoti2020calibrating}, differentiable softmax \citet{KarandikarSoftmax}, etc. (see \cite{minderer2021revisiting} for a recap of related calibration techniques for neural networks). For neural networks, we . This has not been as well-studied in boosting frameworks. However, \cite{Gruber2022} shows that the Brier score is a reasonable choice of loss function for achieving calibration (defined across a variety of CE metrics). 
            \item \textbf{Multicalibration} (Post): Multicalibration \cite{pmlr-v80-hebert-johnson18a} and its variant Multiaccuracy \cite{DBLP:journals/corr/abs-1805-12317} have shown promises on improving multi-group fairness by iteratively calibrating subpopulations that are defined by the intersections of features. Presumably, it works better when there exist features which correlate with the protected attribute. We implemented various options of this approach as hyperparameters to tune (see \ref{App: Multicalibration} for details), and selected the best option to test. 
            \item \textbf{General calibrators} (Post):  Post-hoc calibrators apply a confidence mapping $g$ on top of a mis-calibrated scoring classifier $\hat{p} = h(x)$ to deliver a calibrated confidence score $\hat{q} = g(h(x))$, to improve (average-case) calibration on the \textit{overall} population \cite{Kueppers_2020_CVPR_Workshops}. This is in contrast to the per-group calibrator mentioned above which customizes the calibration for each group. These methods include histogram calibration, isotonic regression \cite{MizilPlattIsotonic}, Platt scaling \cite{Platt99probabilisticoutputs}, Bayesian Binning into Quantiles (BBQ) \cite{naeini2015bbq}, Ensemble of Near-Isotonic Regressions (ENIR) \cite{naeini2016ENIR}, Beta calibration \cite{kull2019betaScaling}, and the Platt scaling and histogram binning hybrid as developed in \cite{Kumar2019}. Please see \ref{App: Calibration Methods} for detailed descriptions. We test all of the above calibrators based on the implementations made available by in the Netcal Python package \cite{Kueppers_2020_CVPR_Workshops}.
        \end{itemize}
\end{itemize}

\paragraph{Feature Selection Methods for Predictive Parity} 
In the intervention methods discussed above, we have excluded examining the highly practical choice of making the model better by finding more/better features, which has been explored in literature \cite{CaiFairAllocation, Ghosh_2022}. One may even argue that adding features linked to group data presents a theoretically backed strategy towards PP fairness based on the findings of \citet{LiuImplicitFairnessCriterion}. However, as mentioned in Section \ref{Section: Framework} in the context of engineering efforts, we find that most large companies pursing AI fairness operate via a centralized responsible AI team. This means that finding features for each model that both improve performance and fairness can be greatly labor intensive and difficult to scale across many models. 

Yet, in reality where scaling up responsible AI efforts is not a priority (e.g., only a few consumer-facing models need to be made fairer), this is a reasonable option to pursue and we also present a new feature-selection based method towards PP fairness. In many scenarios, we often have other non-demographic data used in other production systems in the organization but are not included in a particular model due to being deemed as irrelevant or was phased out in the feature selection step. Some of such features, though it may not aid in accuracy, may improve fairness. Given that including all features is often not realistic (due high cost of collection or pipeline maintenance) we aim to select the right feature(s) to use in our model. We propose a quantile-based bias metric, Quantile Prediction Drift (QPD), which is derived from the Quantile Demographic Drift (QDD) bias metric proposed by \citet{Ghosh_2022}. We show that QPD feature attributions is an efficient way to help find features that could help achieve PP if included in the model. Due to the greater dependence on context for this methodology, we leave the details in Appendix \ref{App: Methodology list} and experimental results in \ref{App: Results for ECCE}.

%% file: Sections/CaseStudy.tex
\section{Case Study} \label{section: Case Study}

Having set up the fairness metric of interest (predictive parity), defined its measurement (worst-group calibration error), and established several methodologies to benchmark, we now move on to our case study. To first summarize, our experiments will show the following results:
\begin{enumerate}
    \item \textbf{Usage of group data appears to be necessary for achieving PP fairness}. We find that methods that do not use any group data (e.g., average-case calibrators and multicalibration) do not provide much benefit over the baseline of tuning for accuracy. 
    \item \textbf{Group-robust optimization in training offers a promising path toward PP fairness}. We find that our proposed DRO-based methods are consistently Pareto optimal in the sense that no method in our benchmarking can achieve better fairness without requiring more group data. These findings point to further research opportunities that to our knowledge have not been explored for PP fairness.
    \item \textbf{Model inductive bias is a key part of fairness}. We find that XGBoost consistently outperforms NN on all fronts (accuracy, fairness, ease of implementation) on tabular data. This emphasizes that the base model choice may have significant impacts on fairness properties and thus should not be overlooked. 
\end{enumerate}

\subsection{Overview}

Our proposition for operationalizing fairness interventions is to quantitatively and qualitatively consider the trade-offs between four pillars of (1) data availability (2) place of intervention and engineering effort (3) intervention effectiveness (4) model performance. Of these, we find (1), (3), and (4) to be the most quantifiable and demonstrate our framework by quantifying the trade-offs between these factors. In applying our framework in real situations, engineering effort would have to be considered but given that its quantification is organization and team-dependent, we leave it out of the evaluation for now. 

We focus our testing on methods to achieve predictive parity through calibration in binary classification settings with tabular data. The overarching process for these experiments is as follows and for each data set considered, we run these steps over three trials to produce means and standard deviations for all reported metrics:
\begin{enumerate}
    \item Split the data randomly into training-validation-testing data sets using a 60/20/20 split. 
    \item Tune a base model (XGB or neural network) by tuning hyperparameters over a fixed grid to optimize for a particular metric. This is done by fitting on the training data and evaluating hyperparameter quality on the validation data. 
    \item Use the best set of hyperparameters found in the previous step to fit the base model on the training data. 
    \item Apply the predictive parity mitigation technique(s). 
    \item Predict and evaluate the effectiveness of the mitigation method(s) on the testing data.
\end{enumerate}

\subsection{Experiment results}

\begin{center}
    \begin{figure}[h]
    \caption{The Pareto frontier of fairness interventions for two model classes (XGBoost and Neural Networks) aggregating results over the six ACS datasets. The degree of predictive parity fairness is plotted on the Y-axis and data requirement on the X-axis. Each point represents a unique fairness intervention method for predictive parity (points close to the bottom left-hand corner are better). We have labeled the Pareto optimal points as well as the baseline points "Tune for Acc." Model performance is considered by observing the accuracy range.}
    \label{fig:Aggregate ACS Frontier}
    \includegraphics[width=\textwidth]{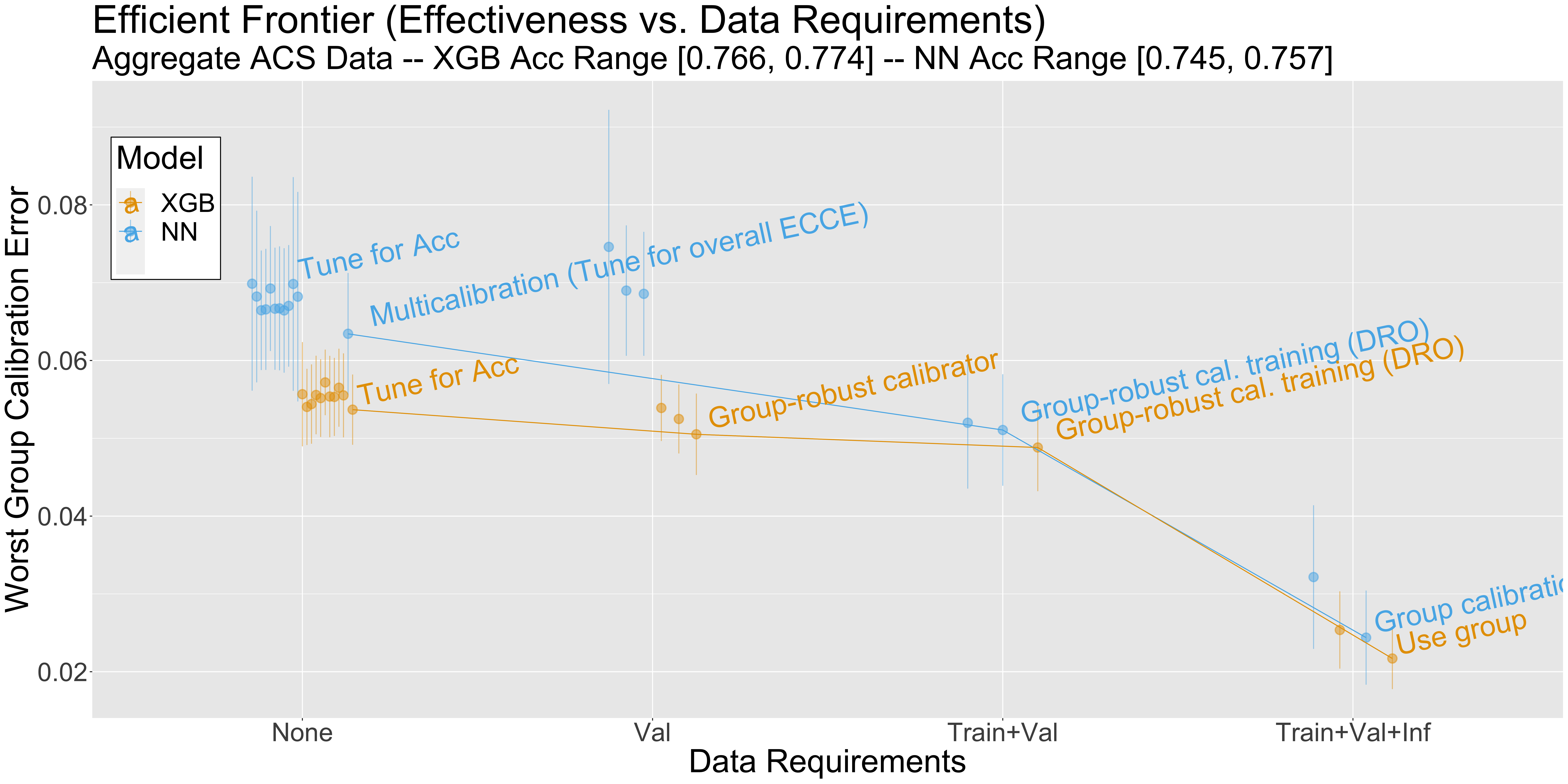}
    \end{figure}
\end{center}

Through our experiments, we can produce a Pareto frontier that depicts the trade-offs between data requirements and intervention effectiveness. We first illustrate this frontier when aggregating results across the six ACS dataset\footnote{By aggregation, we define the point values as the mean across the accuracy and worst-group calibration error across datasets. Similarly, confidence intervals are computed as the mean of the standard deviations for the respective statistics} to highlight some overarching themes (Figure \ref{fig:Aggregate ACS Frontier}).  Then, we will aggregate the results across all ten datasets to draw conclusions. We will report equivalent statistics when defining bias on the basis of worst-case MSCE and KCE in the appendix, but for now note that the results are generally aligned (especially between ECCE and MSCE). 

Looking at Figure \ref{fig:Aggregate ACS Frontier} across the two model classes first, we observe that the Pareto frontier for the XGB model lies entirely below the corresponding line and furthermore the minimum overall accuracy (shown in the subtitle of Figure \ref{fig:Aggregate ACS Frontier}) across all methods for XGB is higher than the maximum overall accuracy than NN models. This demonstrates the importance of the inductive biases and reinforces the findings of \cite{GardnerXGB}, which underscores the extremely hard-to-beat performance of XGBoost on tabular data regardless of what we are measuring. 

Next, we look at similarities across both models. We first expectantly observe that more data availability tends to yield lower bias, which highlights the importance of our data requirement pillar. We also observe that despite the wide array of interventions attempted, the overall accuracy does not vary by much. This is a known aspect of predictive parity fairness, whereby unlike other notions such as demographic parity/equalized odds, inducing predictive parity has relatively muted effects or can even improve metrics like accuracy or ROC AUC  \cite{DurfeeHetCal}. More interestingly, we observe that some intuitive solutions for addressing predictive parity do not work as well as one may initially expect. For instance, while it may be intuitive that a post-hoc calibrator such as isotonic regression or Platt scaling may achieve much better worst-group calibration, it turns out these methods are not much better than optimizing for accuracy. This may be due to the fact that the worst-case performance is often observed in the smallest group and thus while calibrators may shrink overall calibration error, it does little to no favors for the worst group and thus fails to achieve PP fairness.  

Next, we will rely on notions of Pareto optimality to aggregate the results over our 10 datasets (Tables \ref{tbl: XGBsummary}, \ref{tbl: NNsummary}) and distill them into a set of conclusions. For each dataset, we will first identify the best method in each data availability category (None, Val, Train+Val, or Train+Val+Inference) and determine if it is Pareto optimal (with respect to the multi-objective optimization problem of minimizing worst-group calibration error and minimizing use of group attributes). Then, we will look across the 10 datasets and count the number of times a category produced a Pareto optimal method (third column of Tables \ref{tbl: XGBsummary}, \ref{tbl: NNsummary}). Finally, conditional on a data availability category having a Pareto optimal method, we report the most frequently optimal method across all such instances (fourth column of Tables \ref{tbl: XGBsummary}, \ref{tbl: NNsummary})\footnote{To give an example of how this table can be read, row 3 of Table \ref{tbl: XGBsummary} says that in 7 out of 10 datasets, getting lower worst-group calibration error than the best method of Val data  requires using more group data. Of those 7 instances, the group-robust calibrator method is the best perfoming in 4/7 of them.}. 

\begin{center}
\begin{table}[h]
\begin{tabulary}{\linewidth}{||LCCL||} 
 \hline
 Data Availability & \# Methods Tested & \# Times Optimal Across Datasets & Best Pareto Optimal Method \\ [0.5ex] 
 \hline\hline
 Train+Val+Inf & 2 & 7 & Use group (5/7) \\ 
 \hline
 Train+Val & 1 & 3 & Group-robust calibration training (3/3) \\
 \hline
 Val & 3 & 7 & Group-robust calibrator (4/7) \\
 \hline
 None & 10 & 10 & Tune for accuracy (3/10) \\ [1ex] 
 \hline
\end{tabulary}
 
\caption{Summarizing benchmarks across 10 datasets for XGB}
\label{tbl: XGBsummary}
\end{table}
\end{center}

\vspace{-1cm}

\begin{center}
\begin{table}[h]
\begin{tabulary}{\linewidth}{||LCCL||} 
 \hline
 Data Availability & \# Methods Tested & \# Times Optimal Across Datasets & Best Pareto Optimal Method \\ [0.5ex] 
 \hline\hline
 Train+Val+Inf & 2 & 8 & Use group (4/8) \\ 
 \hline
 Train+Val & 2 & 7 & Group-robust calibration training (4/7) \\
 \hline
 Val & 3 & 1 & Group-robust calibrator (1/1) \\
 \hline
 None & 12 & 10 & Multicalibration (4/10) \\ [1ex] 
 \hline
\end{tabulary}
 
\caption{Summarizing benchmarks across 10 datasets for NN}
\label{tbl: NNsummary}  
\end{table}
\end{center}

We now synthesize these results into conclusions and recommendations for practitioners. 

\paragraph{\textbf{Usage of group data appears to be necessary for achieving PP fairness}}
Focusing first on the two extremes - full availability and no availability, the clearest result is that the most consistent and effective way to achieve PP fairness is to use the group attribute as a feature (though whether this \textit{should} be done is a separate discussion which we will return to in the conclusion). This result is not surprising, but it does hint that theoretical results showing that PP fairness comes for free under sufficiently rich models/features \cite{LiuImplicitFairnessCriterion} should not be overly relied on in practice, as there are several commonly used models (XGB, neural networks) that still benefit largely from the use of group data or otherwise situations where features are not informative enough. On the other end of the spectrum, when no group data is available, we find that average-case calibrators and even multicalibration cannot consistently beat the baseline of simply tuning for accuracy. This is likely due to the fact that average-case calibrators mostly improve calibration for the largest group, which is typically not the group with the worst-case calibration error. We summarize these observations in the following remark.

\begin{remark}
    The lack of empirical effectiveness across several mitigation strategies that avoid using group data indicates that practitioners seeking to achieve PP may need at least some amount of group data to reliably fix gaps in predictive parity.
\end{remark}

\paragraph{\textbf{Group-robust calibration offers a promising path toward PP fairness}}
Turning to the results in the val and train+val data availability regimes, we find that our proposed methods of using a group-robust calibrator or group-robust calibration training show strong promise. In particular, the group-robust calibration training (\ref{eq: droCalLoss}) works consistently well in achieving PP fairness for the neural network model\footnote{The method (or the method with temperature scaling) was Pareto optimal in 6/6 ACS datasets, but did not work as well in the small datasets: German, Bank, COMPAS, perhaps due to the lack of signal in the smaller data.}. This success is likely due to the effectiveness of group DRO \cite{sagawa2019distributionally} and MMCE \cite{kumar2018trainable, BlasiokCalibration} being a theoretically-backed measurement of calibration error. The equivalent heuristic technique for XGBoost was not as effective likely due to having a less theoretically principled implementation. However, Pareto dominance on 3 datasets indicates that customizing losses in XGBoost to focus on group-robust calibration loss deserves further investigation as an in-processing mitigation approach. Overall, we emphasize that these are techniques that effectively improve PP fairness \emph{without using group data at inference}.

\paragraph{\textbf{Model inductive bias is a key part of fairness}}
As shown in Figure \ref{fig:Aggregate ACS Frontier}, the frontier of XGBoost sits below that of the NN model and furthermore the best accuracy of the NN models is lower than the worst accuracy across the XGBoost models. To top it off, we found that XGBoost was much easier to work with (in terms of training and tuning) while setting up these large-scale experiments. This is in line with a recent study of subgroup fairness and robustness properties of XGBoost models \cite{GardnerXGB} in which the authors found that on tabular data, XGB tuned for accuracy is a very hard-to-beat baseline (though the study did not evaluate for PP fairness). 

%% file: Sections/Conclusion.tex
\section{Conclusion}

Recent literature has made it clear that AI fairness is a critical problem that absolutely requires collaboration between AI fairness practitioners as well as researchers on both the algorithmic and social/legal theory. Our works overall aims to help bridge this collaboration more closely by presenting a holistic framework that captures social/legal factors (use of group data) and industrial factors (engineering cost/scalability) on top of canonical pillars of algorithmic fairness and performance. Our framework emphasizes that the trade-offs between these four pillars can be quantitatively evaluated by surveying the landscape of fairness algorithms and further considering the data requirements for each algorithm rather than just the standard pre-/in-/post-processing categorization. As we demonstrate with a case study on predictive parity fairness, we hope that illustrating these trade-offs can further fill the gap in communications between system-level directives and researchers on the algorithms side. Our case study also yields methodological findings on achieving predictive parity fairness by shedding light on the need for use of group data, the effectiveness of group-robust calibration training, and the role of model inductive bias in fairness.

Beyond our contributions, we hope that this work adds to the growing movement of practical considerations for fairness. Throughout the legal and social literature that we have reviewed (\cite{AndrusDemographic, AndrusSpitzer, BogenAwareness, bent2019algorithmic, xiang2020reconciling}), the only consistent theme is that there does not seem to be a hard and fast rule for the collection and usage of group data. We therefore urge academics and practitioners to continually monitor the developments of AI fairness in society, law, and industry to research and implement solutions accordingly.

%% file: Sections/Appendix.tex
\section{Appendix}

\subsection{Recap of Recent Calibration Error Literature} \label{App: CE Metrics}
Calibration error metrics measure the calibration of the model (i.e., how closely the confidence of the model corresponds to the accuracy). While this appears to be a straightforward statistic to estimate, there are several ways to define and measure this notion and recent literature has found that not all methods are made equal. These nuances are important to clarify for the measurement of predictive parity as we have defined it based on the worst-group calibration error (refer to Section \ref{subsection: PP and Cal} for our reasoning). We now dive into these details by recapping research into calibration error metrics and then justifying our choice of using the Expected Cumulative Calibration Error (ECCE). 

We denote the calibration of a measurement function $f$ as $CE(f)$. One of the most widely used measures of calibration error is the Expected Calibration Error (ECE) as proposed in \cite{naeini2015bbq} and widely popularized in \cite{guoCalibration}. ECE aims to capture the discrepancy between confidence and accuracy. Intuitively, ECE is computed by first splitting up the model score range into $M$ evenly spaced bins (e.g., [0, 0.1), [0.1, 0.2), etc.), then computing the expectation in outcome $Y_{m}$ in each bin (known as the bin accuracy $acc(B_{m})$) and the expected score $s$ (also known as the bin confidence $conf(B_{m})$) and taking the difference of the two terms. In terms of actual formulas, this is captured as: 
\begin{equation}
    ECE = \sum_{m=1}^{M} \frac{\lvert B_{m} \rvert}{n} \lvert acc(B_{m})-conf(B_{m}) \rvert \label{eq: ECE}
\end{equation}

Where $M$ corresponds to an a-priori selected number of bins to evaluate and $n$ is the total number of samples.  However, several recent works have observed theoretical and empirical deficiencies with this metric \cite{Kumar2019, roelofs2022mitigating, Gruber2022, BlasiokCalibration, arrieta2022metrics}. To summarize these criticisms, ECE is a statistically biased (under-estimator) measure of true calibration error due to the need to parameterize the measurement with a a-priori fixed number of $M$ bins. The bias is worsened by ECE's sensitivity to sample size \cite{roelofs2022mitigating, Gruber2022}. The aforementioned works all propose alternative CE measurements that demonstrate better empirical performance \cite{roelofs2022mitigating} or have stricter theoretical requirements and properties such having a its distance to some true calibration error (TCE) be polynomially bounded (e.g., $CE(f) \leq \mathcal{O}(TCE^{s})$ for some constant $s$ \cite{BlasiokCalibration}) or have asymptotic convergence properties to the true calibration error \cite{arrieta2022metrics}. After evaluating the recent landscape of CE metrics, we have chosen to actively avoid using ECE and instead focus on three such metrics, described below in order of preference with accompanying justification. We recommend \cite{Gruber2022} and \cite{BlasiokCalibration} for a thorough survey of the CE landscape.

\begin{itemize}
  \item \textbf{Expected Cumulative Calibration Error} (ECCE): Proposed in \cite{arrieta2022metrics}, this metric offers a theoretically-backed, parameter-free, and scalable CE metric that enables hypothesis testing. The crux of this metric is that instead of using notions related to binning, they use the cumulative calibration error which presents asymptotic properties of converging to zero for a perfectly calibrated set of scores and being bounded away from zero for uncalibrated scores. Across our experiments, we rely on this CE the most to draw conclusions as it is has both theoretical properties and empirically works well. \footnote{Based on experiments following the bias-by-construction framework as presented in \cite{roelofs2022mitigating}}
  \item \textbf{Minimum Sweep Calibration Error} (MSCE): Proposed in \cite{roelofs2022mitigating}, this metric is a modification of the equal-mass binning version of the original ECE~\eqref{eq: ECE} where the number of bins is dynamically selected. The authors also validate its empirical effectiveness over ECE through the proposed bias-by-construction framework, where the $CE(f)=E_{D}[CE(f)]-TCE$ where $D$ is the distribution of scores and labels and $TCE$ is the true calibration error, which has a parametric form and thus can be numerically approximated. We favor this CE due to its empirical effectiveness and empirical alignment with the traditional ECE metric. 
  \item \textbf{Kernel Calibration Error} (KCE): Proposed in \cite{kumar2018trainable} and further analyzed in \cite{BlasiokCalibration}, this metric is a reproducing kernel Hilbert space (RKHS) based measurement of calibration. Originally designed to be a differentiable and consistent estimator of ECE, Błasiok et al. show that it also carries theoretical properties of being a polynomial upper and lower bound to a notion of true calibration error \cite{BlasiokCalibration}. Although this CE has strong theoretical properties, it is expensive to compute without sampling as it requires $\mathcal{O}(n^{2})$ computations. As noted in \cite{BlasiokCalibration}, a practical implementation can bypass the $n^{2}$ scaling by drawing $M$ samples instead and relying on the error shrinkage rate of $\mathcal{O}(\frac{1}{\sqrt{M}})$. However, this results in another practical problem in that the measurement is stochastic. For these reasons, we include KCE, but do not rely strongly on its results to draw conclusions.
\end{itemize}

We acknowledge that this is not a comprehensive list of proposed calibration metrics and we have evaluated metrics such as the Brier score (suggested and highlighted as a proper score in \cite{Gruber2022}) and kernel estimation as proposed in \cite{diciccio2022predictive}, but have omitted them for simplicity as it is difficult to pick a universally "best" CE metric given the current literature. We hope that our work incites further discussion for fairness intervention methods for predictive parity and as a side-consequence, we hope that it encourages the academic community to converge on a definition for calibration error. 

\subsection{Dataset Descriptions}
We describe the datasets we use in the table below. For the ACS Datasets, we use the API provided in \citet{DingRetiring}. This API allows us to use the standard features for each modeling task. For group encoding, we keep the original encoding except that we group the smallest Hawaiian islander/Native American/Indian groups together to avoid minuscule group sizes. We use the 2018, 1-person data collected over the states CA, OR, WA, ID, AK, HI, NV, UT, MT, WY, CO, NM. 

We also benchmark on some of the most common fairness benchmarking datasets as surveyed by \citet{HortMitigationSurvey}. This includes the Bank, German, and Adult dataset as featured on the UCI Data Repository \cite{DuaUCI}. We also gathered COMPAS data \cite{COMPAS_angwin_larson_mattu_kirchner_2016} as it is featured on Github\footnote{https://github.com/propublica/compas-analysis}. 

\begin{table}[h]

\begin{tabular}{llllll}
\toprule
Dataset & Training N & N Groups & N of Smallest Group & Smallest Group E[Y] & Other Groups E[Y] \\
\midrule
ACS Travel Time &     145569 &        6 &                4965 &            0.451158 &          0.333333 \\
ACS Mobility &      84429 &        6 &                2460 &             0.80813 &           0.73196 \\
ACS Income &     214054 &        6 &                4879 &             0.22894 &          0.398212 \\
ACS Income Poverty &     412056 &        6 &               10948 &            0.543752 &          0.330103 \\
ACS Public Coverage &     145569 &        6 &                4965 &            0.451158 &          0.333333 \\
ACS Employment &     412056 &        6 &               10948 &            0.374132 &          0.459938 \\
Bank &      24712 &        2 &                 617 &            0.231767 &          0.110687 \\
 COMPAS &       4328 &        6 &                   6 &            0.833333 &          0.446784 \\
German &        600 &        3 &                  87 &            0.333333 &          0.290448 \\
Adult &      29304 &        5 &                 246 &            0.142276 &          0.239246 \\
\bottomrule
\end{tabular}
\caption{Description of the datasets for each task that we model.}
\label{tbl: datasets}
\end{table}

\subsection{Detailed Description of Experimental Setup} \label{App: Experimental setup}
We used PyTorch \cite{pytorch} for the neural network implementations and the DMLC XGBoost implementation for the boosting model \cite{Chen:2016:XST:2939672.2939785}. For neural networks, we stick to only 2-layer networks (with batch normalization, ReLu activation, and L2 normalization), trained by the stochastic gradient descent (SGD) optimizer for training/tuning simplicity. For both models, numerical data is used as-is and all categorical variables are one-hot encoded into one or more features.

For each methodological combination, we use the Ray Tune \cite{liaw2018tune} paradigm to tune our models over 20 trials and optimize for either overall accuracy, overall ECCE (in regimes where group data is unavailable), or worst-group ECCE (if group data is available in validation). In doing so, we set the following hyperparameter search spaces (parameters with asterisk are default search points): 
\begin{center}
\begin{table}
\begin{tabulary}{1.0\textwidth}{|L|L|L|L|}
\hline
\textbf{Method} & \textbf{Tuning Parameters} & \textbf{Values} \\
\hline
\multirow{3}{15em}{XGBoost} & Eta (learning rate) & \{0.1, 0.3*, 1.0\} \\ 
& Min split loss & \{0.0*, 0.1, 0.5\} \\ 
& Max depth &  \{4, 6*, 8\}\\ 
& Colsample by tree & \{0.7, 0.9, 1.0*\} \\ 
& Colsample by level &  \{0.7, 0.9, 1.0*\} \\ 
& Max bin & \{ 128, 256, 512*\}\\ 
& Grow policy &  \{depthwise, lossguide*\}\\ 
& Boosting rounds &  \{10, 25*, 50, 100\}\\ 
\hline
\multirow{3}{15em}{XGBoost (Calibration loss training)} & Eta (learning rate) & \{0.1, 0.3*, 1.0\} \\ 
& Min split loss & \{0.0*, 0.1, 0.5\} \\ 
& Max depth &  \{4, 6*, 8\}\\ 
& Colsample by tree & \{0.7, 0.9, 1.0*\} \\ 
& Colsample by level &  \{0.7, 0.9, 1.0*\} \\ 
& Max bin & \{ 128, 256, 512*\}\\ 
& Grow policy &  \{depthwise, lossguide*\}\\ 
& Boosting rounds &  \{10, 25*, 50, 100\}\\ 
& Calibration loss weight & \{1e-3, 1e-2, 1e-1*, 1, 2\}\\
\hline
\multirow{3}{15em}{XGBoost (Group-robust calibration training)} & Eta (learning rate) & \{0.1, 0.3*, 1.0\} \\ 
& Min split loss & \{0.0*, 0.1, 0.5\} \\ 
& Max depth &  \{4, 6*, 8\}\\ 
& Colsample by tree & \{0.7, 0.9, 1.0*\} \\ 
& Colsample by level &  \{0.7, 0.9, 1.0*\} \\ 
& Max bin & \{ 128, 256, 512*\}\\ 
& Grow policy &  \{depthwise, lossguide*\}\\ 
& Boosting rounds &  \{10, 25*, 50, 100\}\\ 
& Calibration loss weight & \{1e-3, 1e-2, 1e-1*, 1, 2\}\\
& DRO Eta & \{1.0, 2.0*, 5.0, 10.0\}\\
\hline
\multirow{3}{15em}{Neural Network} & Layer 1 units & \{16, 32, 64, 128*, 256, 512\} \\ 
& Layer 2 units & \{16, 32, 64*, 128, 256, 512\}  \\ 
& Learning rate & Loguniform \{1e-4, 1e-3*,1e-2\} \\ 
& L2-Regularization &  Loguniform \{1e-6, 1e-5*, 1e-2\}\\ 
& Batch size & \{64, 128, 512*, 1024\} \\ 
& Num epochs &  \{10, 20, 30*, 40, 50\} \\ 
\hline
\multirow{3}{15em}{Neural Network (Calibration loss training)} & Layer 1 units & \{16, 32, 64, 128*, 256, 512\} \\ 
& Layer 2 units & \{16, 32, 64*, 128, 256, 512\}  \\ 
& Learning rate & Loguniform \{1e-4, 1e-3*,1e-2\} \\ 
& L2-Regularization &  Loguniform \{1e-6, 1e-5*, 1e-2\}\\ 
& Batch size & \{64, 128, 512*, 1024\} \\ 
& Num epochs &  \{10, 20, 30*, 40, 50\} \\ 
& Calibration Loss Weight & Loguniform \{1e-3, 1e-2*, 1.0\} \\
\hline
\multirow{3}{15em}{Neural Network (Group-robust calibration training)} & Layer 1 units & \{16, 32, 64, 128*, 256, 512\} \\ 
& Layer 2 units & \{16, 32, 64*, 128, 256, 512\}  \\ 
& Learning rate & Loguniform \{1e-4, 1e-3*,1e-2\} \\ 
& L2-Regularization &  Loguniform \{1e-6, 1e-5*, 1e-2\}\\ 
& Batch size & \{64, 128, 512*, 1024\} \\ 
& Num epochs &  \{10, 20, 30*, 40, 50\} \\ 
& Calibration Loss Weight & Loguniform \{1e-3, 1e-2*, 1.0\} \\
& DRO Eta & Loguniform \{1e-4, 1e-3*, 1.0\} \\
& DRO Regularization & Loguniform \{1e-3, 1e-2*, 1.0\} \\
\hline
\end{tabulary}
\end{table}
\end{center}

\subsection{Detailed Description of Methodologies} \label{App: Methodology list}
In this section, we will provide more in-depth descriptions about the methodologies discussed in Section \ref{section: Case Study}. We will do so in order of how they are presented, omitting certain methodologies as we felt that they are sufficiently described in the main section (e.g., group as a feature, per-group calibrator, tune for worst group calibration, tune for accuracy). 

\subsubsection{Group-robust calibration training}
For the neural network implementation, the group-robust calibration training scheme optimizes the following loss function. 
\begin{equation}
    \mathcal{L}(x,y,g)=\sum_{i} y_{i}log(f(x_{i}))+(1-y_{i})log(1-f(x_{i}))+\gamma( \underset{g\in\mathcal{G}}{max}E_{(x,y)}\sim\hat{P_{g}}[CE(f)])
\end{equation} 
The intuition behind this loss function is that we have a basic loss (binary cross-entropy) that is typically used for binary classification tasks and it is combined with a DRO-style loss that tells the model to focus more on groups where the loss is greater. Formal details (including implementation) of the Group DRO methodology can be found in \citet{sagawa2019distributionally}. The $CE(f)$ term was chosen to be the MMCE kernel as proposed in \citet{kumar2018trainable} and further theoretically analyzed in \citet{BlasiokCalibration}. We repeat it below for completeness:
\begin{equation}
    MMCE^{2}_{m}=\sum_{i,j\in \mathcal(D)}\frac{(c_{i}-r_{i})(c_{j}-r_{j})k(r_{i},r_{j})}{m^{2}} \label{eq: MMCE}
\end{equation}
where $m$ is the number of samples in a group batch and $i,j$ are samples with $c_{i}$ is the correctness score (equals 1 if $\hat{y_{i}}=y_{i}$ and 0 otherwise) and $r_{i}$ is the model outputted score of the sample. The authors show that theoretically, this function is equal to zero only for perfectly calibrated models and intuitively, it applies a dampening force to overconfident predictions that approach 1. It is worth noting that we also attempted other loss functions in our experiments, including focal loss \cite{mukhoti2020calibrating} and Brier score (inspired by the findings in \citet{Gruber2022} which motivates the use of Brier score). However, we found that the that we used worked the most consistently given our experimental setup. 

For the XGBoost implementation, we pivoted to using a Brier score loss as the regularization term. The Brier score is also a theoretically backed measure of calibration as it is decomposable into refinement and calibration as a proper score \cite{murphy1986new, Gruber2022}. The reason we use the Brier score instead of MMCE is simply due to the ease of implementation, as XGBoost does not feature automatic gradient/Hessian computations unlike the Torch framework. Brier score has a easily expressible gradient and Hessian and we use these to weigh the samples in each boosting iteration based on the per-group loss relative to the total loss. 

\subsubsection{Feature Selection Based On QPD Attribution}
We propose a quantile-based bias metric, Quantile Prediction Drift (QPD), and discuss how to calculate and leverage the feature importance based on QPD to identify which feature(s) could help reduce bias.

A similar metric, Quantile Demographic Drift (QDD), was proposed by \cite{Ghosh_2022}, which uses the model score difference between groups across different score quantiles. The authors proposed that the quantile-based differences of the feature attributions between groups could be used as fairness explanations. Formally, 
\begin{equation}
\label{eq:qpd0} QDD_b = E_b[S|G=i] - E_b[S|G=j].
\end{equation}
QDD is closely connected with Demographic Parity (DP). Note that DP can be formally stated as  $$E_b[S|G=i] = E_b[S|G=j] \quad \forall i,j \in \{1,\ldots,G | i\neq j \}.$$
We propose QPD, which is a natural derivation of QDD under the context of PP (equation \ref{eq:pp}). Different from QDD, the metric quantifies the difference in expected \textit{outcome} (instead of expected scores) given any score quantile $b$. For any two groups $i$ and $j$, define 
\begin{equation}
\label{eq:qpd1} QPD_b = E[Y | S=s, G=i] - E[Y | S=s, G=j].
\end{equation}
Similar to the derivation in \cite{Ghosh_2022}, QPD can be approximated as 
\begin{equation}
\label{eq:qpd2} QPD_b = \frac{1}{N_i}\sum_{n_i=1}^{N_i}Y_{n_i} - \frac{1}{N_j}\sum_{n_j=1}^{N_j}Y_{n_j},
\end{equation}
and the local QPD attribution for feature $f$ (i.e. the feature importance for feature $f$ for $QPD_b$) using feature attribution method A can be calculated as 
\begin{equation}
\label{eq:qpda0} QPDA_{b,A,f} = \frac{1}{N_i}\sum_{n_i=1}^{N_i}outcomeAttr_{n_i,A,f} - \frac{1}{N_j}\sum_{n_j=1}^{N_j}outcomeAttr_{n_j,A,f},
\end{equation}
where $outcomeAttr_{n_i,A,f}$ refers to the local attribution of feature $f$ for \textit{outcome} $Y_i$ given instance $n_i$, calculated using feature attribution method A.
For an attribution method A (e.g. SHAP) that satisfies the efficiency axiom \cite{Ghosh_2022}, we can show
\begin{equation}
\label{eq:qpda1} \frac{1}{N_i}\sum_{n_i=1}^{N_i}\sum_{f=1}^{F}outcomeAttr_{n_i,A,f} - \frac{1}{N_j}\sum_{n_j=1}^{N_j}\sum_{f=1}^{F}outcomeAttr_{n_j,A,f} = \frac{1}{N_i}\sum_{n_i=1}^{N_i}Y_{n_i} - \frac{1}{N_j}\sum_{n_j=1}^{N_j}Y_{n_j},
\end{equation}
and therefore,
\begin{equation}
\label{eq:qpda2} QPD_b = \sum_{f=1}^{F} QPDA_{b,A,f}.
\end{equation}
Local feature attribution methods (e.g. SHAP) typically calculates the attribution to model scores $S$. In order to calculate the attribution to another random variable $T$, we generally need access to a data generation mechanism $g(X_w)=T_w$, which tells us the ground-truth value of $T$ given the feature values of any hypothetical instance $w$. 
One way to approximate this mechanism is with a superior model. Under the use case of feature selection, we assume that we are able to train a less biased and more accurate model $M_s$ with the additional features, compared to the original model $M_o$, so that $\hat{g} (X_w) = M_s (X_w) = S_{M_s}$ can be a proxy to the outcome $Y_w$. Although the approximation is not perfect, we can use it to identify features which attributes to the bias difference between $M_s$ and $M_o$, and this is more effective when $M_s$ is less biased. 

When we have a large number of available features to be added to the model, it is often more likely that we can build a better model $M_s$ that is closer to the ground truth outcomes; when the number of available features is large, since the alternative methods (e.g. brute-force retraining with all possible feature combinations, or add all features to the model) would be extremely expensive, the feature selection mechanism that we propose would also be more useful.

In our experiments, we identified the 3 (out of 7) worse calibrated demographic groups in terms of ECCE. We defined two groups for QPD attribution: one group contains all instances within the worse calibrated demographic groups, and the other contains all other instances. We calculated QPD attribution using SHAP and the above formulation to identify the importance features related to bias. To mimic realistic scenarios where building and maintaining feature pipelines are expensive, we identify \textit{one} top feature with the highest aggregated QPD attribution to be added to the model.

Based on experiment results, adding the one top feature identified with QPD attribution often has similar effect as adding all features to the model. We conclude that QPD attribution is an efficient way to identify which feature(s) could help achieve better PP.

\subsubsection{Multicalibration} \label{App: Multicalibration}
This methodology was first proposed in 2018 \cite{pmlr-v80-hebert-johnson18a} as a post-processing approach to achieve algorithmic fairness. It has two key assumptions: 1) Bias comes from the model fitting procedure that optimizes for global accuracy, leaving the minority groups under-fitted. However, the training data and labels are unbiased. 2) When the protected attribute is not accessible, there are other correlated features that can jointly identify the subpopulations. The way it works is to iteratively calibrate the original predictor on validation data, partition by partition, until no identifiable subgroup has miscalibration larger than a predefined threshold or the maximum number of iterations has reached. Then the calibrated predictor can be applied to any new dataset with the same features. We evaluated the effectiveness of multicalibration and its variants on reducing worst-group calibration error (ECCE), without using the protected attribute at any stage. 

There are many possible options to configure the exact method of running multicalibration. Instead of hand-picking, we set them as hyperparameters to tune, which allows the best combination to emerge for each use case. Specifically, these hyperparameters fall into the following categories.

(1) Partition strategy: The total number of partitions and how to create them can be tuned. Probability ranges can split evenly between 0 and 1, or split by quantile to have equal number of instances within each partition.

(2) Sampling strategy: At each iteration, without sampling it uses the same data to fit on the residuals. To reduce over-fitting, we could either split the data equally such that every iteration uses a disjoint subset, or perform bootstrapping to introduce some variation.

(3) Training algorithm: Calibration relies on the predicted residuals. At each iteration, we train a ridge regression or decision tree regression model using all the features to predict the remaining residuals. To reduce over-fitting, the regularization strength and the maximum depth of tree can be tuned, respectively.

(4) Probability update strategy: Miscalibration is estimated by the correlation between predicted and observed residuals. At each iteration, only the partition with the largest miscalibration will get an update on the probability scores. Multicalibration \cite{pmlr-v80-hebert-johnson18a} applies an additive weight to the original probability scores, and re-partition the data based on the new probability scores. A slight variant, multiaccuracy \cite{DBLP:journals/corr/abs-1805-12317}, applies a multiplicative weight to the original probability scores, and does not re-partition the data. In addition, the step size for probability update can also be tuned.

(5) Stopping condition: It triggers a stop when the largest miscalibration across all the partitions is smaller than a predefined threshold, or the maximum number of iterations has been reached. Both the threshold and the maximum number of iterations can be tuned.

For each use case, we took 70\% of the validation data to run the algorithm. During the execution, which partition got updated and its corresponding residual model are both stored for all the iterations. This was done for 20 trials with different combinations of hyperparameters. After that, we replayed the algorithm on the rest 30\% of validation data to select the best configuration. The exact version of algorithm was then evaluated on the test dataset. 

\subsubsection{Detailed Description of Calibration Methods} \label{App: Calibration Methods}

The magnitude of the model score can be seen an estimate of confidence in the prediction for classification tasks. Post-hoc calibration methods apply a confidence mapping $g$ on top of a mis-calibrated scoring classifier $\hat{p} = h(x)$ to deliver a calibrated confidence score $\hat{q} = g(h(x))$, in order to reduce or eliminate calibration errors \cite{Kueppers_2020_CVPR_Workshops}.

Below we describe each of the calibration methods that we tested in the experiments:

\begin{itemize}
  \item \textbf{Histogram Binning}: Proposed by \cite{Zadrozny2001ObtainingCP}, histogram binning sorts each prediction into a bin and assigned its calibrated confidence estimate. Specifically, the probability that an example belongs to a class is estimated with the fraction of training examples in the bin that actually belong to the calss. To extend this method to multiclass classification, we apply the method in a 1-vs-all fashion \cite{Kueppers_2020_CVPR_Workshops}.
  \item \textbf{Isotonic Regression}: Isotonic Regression is proposed by \cite{isotonicReg}. This method is similar to Histogram Binning, except that it uses dynamic bin sizes as well as boundaries. It fits a piece-wise constant function to the ground truth labels sorted by given confidence estimates.
  \item \textbf{Bayesian Binning into Quantiles (BBQ)}: Proposed by \cite{naeini2015bbq}, this method utilizes several Histogram Binning instances with different number of bins. It then obtains the calibrated confidence estimate with a weighted sum of these instances. The scoring function is modified per \cite{Kueppers_2020_CVPR_Workshops}.
  \item \textbf{Ensemble of Near Isotonic Regression (ENIR) models}: Proposed by \cite{naeini2016ENIR}, this method allows for a violation of the monotony restrictions, which is different from Isotonic Regression. ENIR uses the modified Pool-Adjacent-Violators Algorithm (mPAVA) and builds multiple Near Isotonic Regression models. Similar to BBQ, it then uses a weighted score function based on these model instances.
  \item \textbf{Logistic Calibration}: Logistic Calibration is also called Platt Scaling. It is proposed by \cite{Platt2007ProbabilisticOF}. This method uses multiple independent normal distributions to obtain a calibration mapping by means of the confidence and additional features. Specifically, $$g(s) = \frac{1}{1+exp(-z(s))}$$ where $z(s)=lr(s)=s^Tw+c$ is a logistic regression. This calibration scheme assumes independence between all variables.
  \item \textbf{Beta Calibration}: Proposed by \cite{Kull2017BetaCA}, this method is similar to Logistic Calibration except for a different $lr(s)$ formulation. It uses multiple independent Beta distributions to obtain a calibration mapping by means of the confidence as well as additional features. This calibration scheme assumes independence between all variables.
  \item \textbf{Temperature Scaling}: This method is proposed by \cite{guoCalibration}. It divides the logits (inputs to the softmax function) by a learned temperature scaling scalar parameter $T$, so that $$\hat{q} = \sigma(\frac{s}{T})$$ where $s=(\hat{p}, \hat{r})$ is a combined input and $\hat{r} \in {[0,1]}^J$ is an additional box regression output. 
  \item \textbf{PlattBinner}: This method is proposed by \cite{Kumar2019} to address the issue that the calibration error of methods like Platt scaling and temperature scaling are typically underestimated and cannot be measured, and the calibration error of methods like Histogram Binning requires many examples to measure. The authors propose that PlattBinner is an efficient recalibration method where the calibration error can be measured with fewer examples.
\end{itemize}

\subsection{Results for ECCE}  \label{App: Results for ECCE}
In this table we show the Pareto frontier for the full sets of data and methodologies. First for the six ACS datasets and afterwards for all ten datasets. We separate the results of the two because we consider ACS data more reliable as the data size is much larger as shown in Table \ref{tbl: datasets}. Importantly, the variance of the methods increase dramatically when we introduce the additional four UCI small datasets.

In addition to the None, Val, Train+Val, Train+Val+Inference split that we showed and discussed in the main discussion, we also show and label the results for the feature selection methods (QPD and add all additional features). We repeat the same table for MSCE in Appendix \ref{App: Results for MSCE} and KCE in Appendix \ref{fig:KCEFrontiers}. 

\begin{center}
    \begin{figure}[h!] 
    \caption{Pareto frontiers when aggregating data across ACS data (top) and All data (bottom). The degree of predictive parity fairness is plotted on the Y-axis and data requirement on the X-axis. Each point represents a unique fairness intervention method for predictive parity (points close to the bottom left-hand corner are better). We have labeled the Pareto optimal points as well as the baseline points "Tune for Acc." Model performance is considered by observing the accuracy range. We have also added the "Other" category for QPD, but do not consider it to be part of the Pareto frontier.}
    \label{fig:ECCEFrontiers}
    \includegraphics[width=\textwidth]{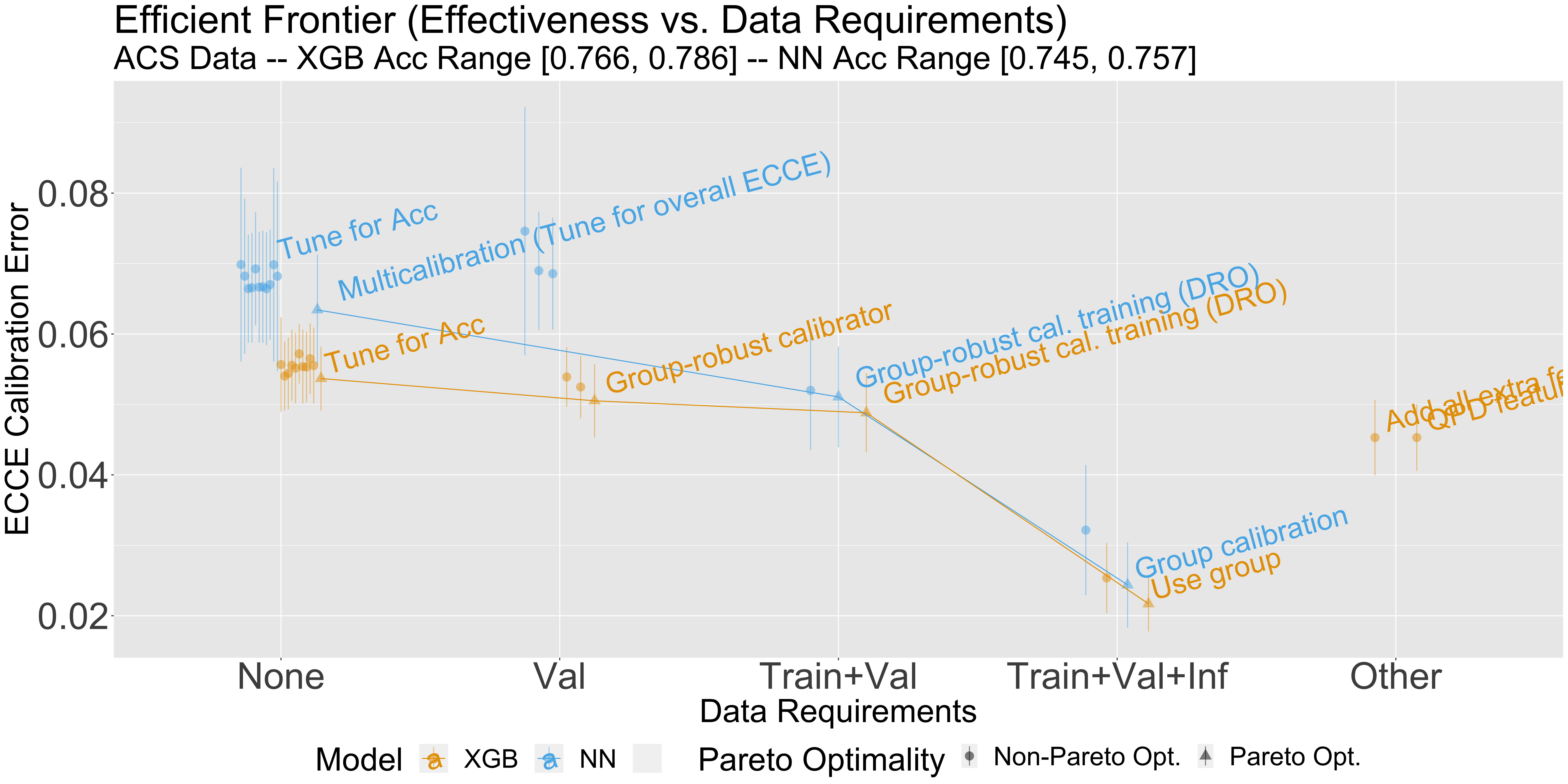}
    \includegraphics[width=\textwidth]{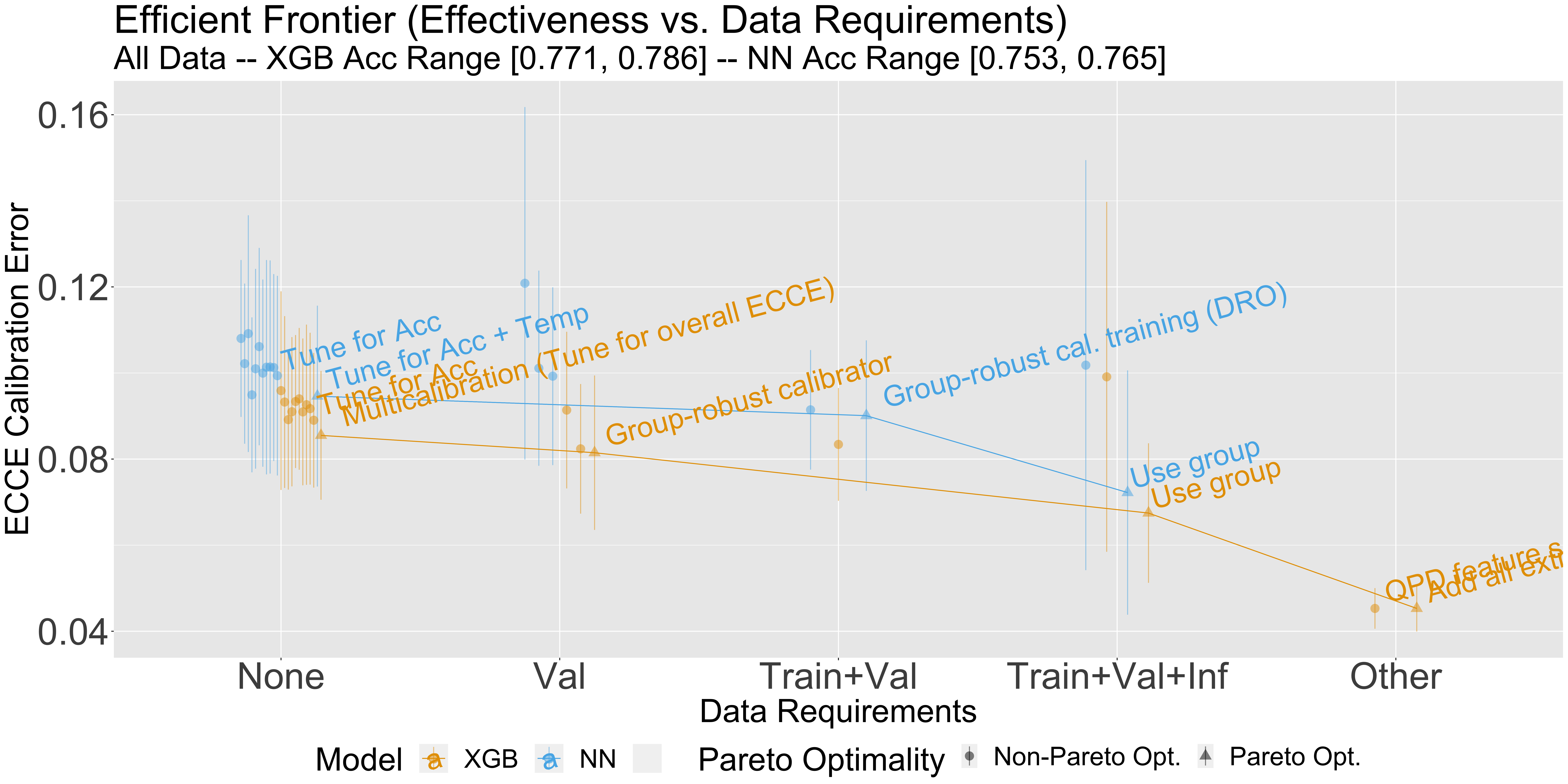}
    \end{figure}
\end{center}

\subsection{Results for MSCE}  \label{App: Results for MSCE}
\begin{center}
    \begin{figure}[h!] 
    \caption{Pareto frontiers when aggregating data across ACS data (top) and All data (bottom). The degree of predictive parity fairness is plotted on the Y-axis and data requirement on the X-axis. Each point represents a unique fairness intervention method for predictive parity (points close to the bottom left-hand corner are better). We have labeled the Pareto optimal points as well as the baseline points "Tune for Acc." Model performance is considered by observing the accuracy range. We have also added the "Other" category for QPD, but do not consider it to be part of the Pareto frontier.}
    \label{fig:MSCEFrontiers}
    \includegraphics[width=\textwidth]{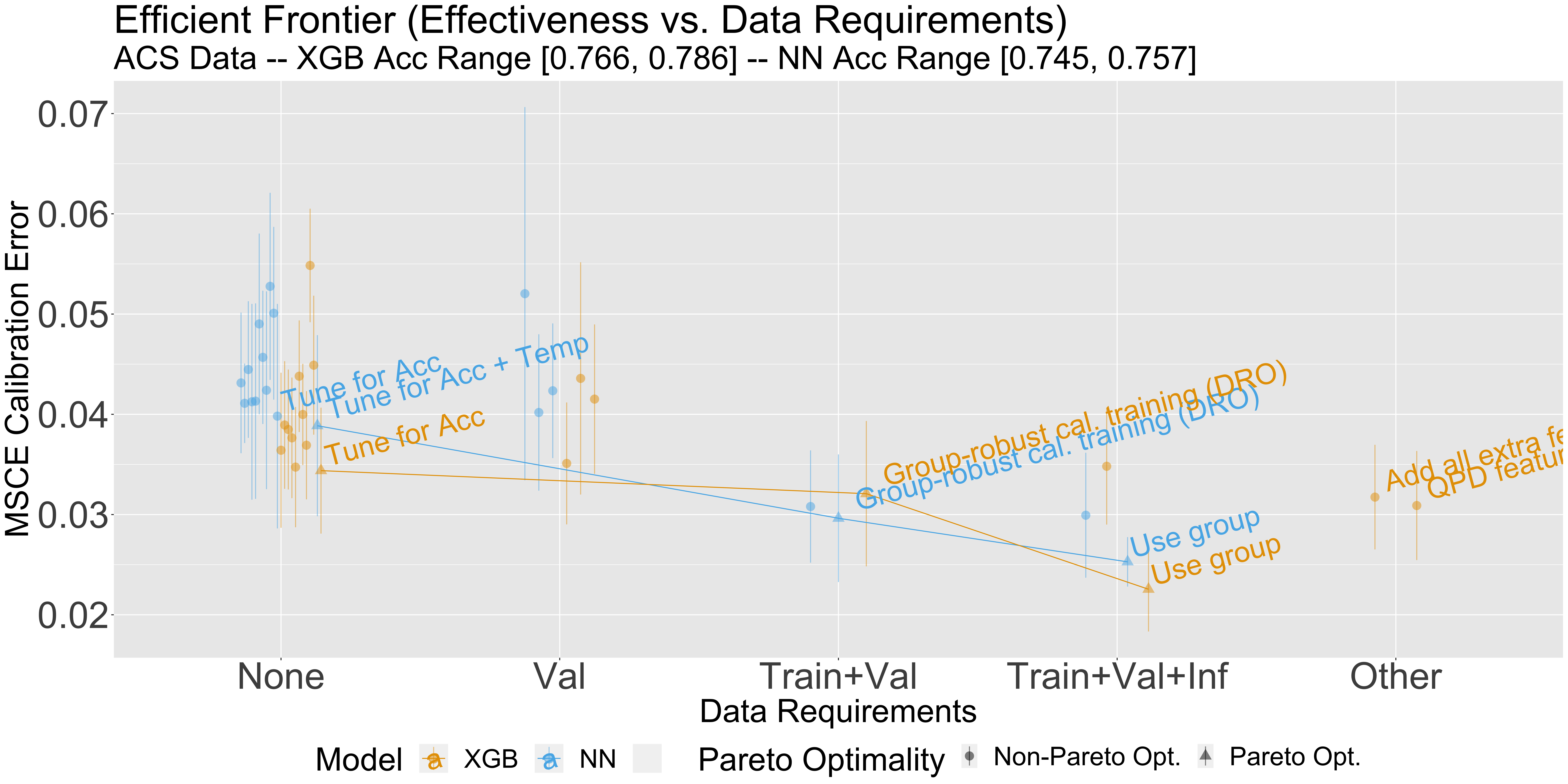}
    \includegraphics[width=\textwidth]{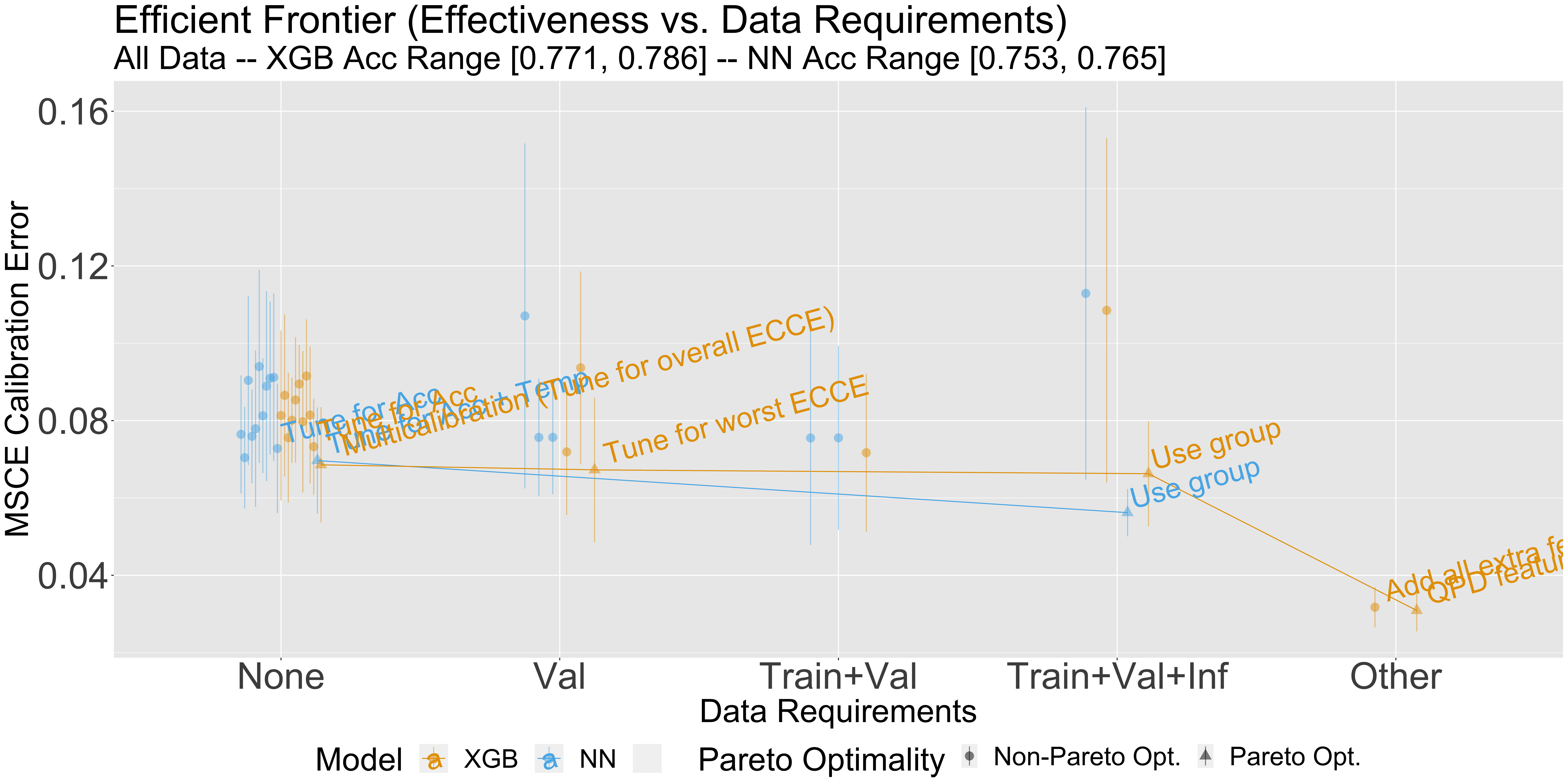}
    \end{figure}
\end{center}

Our study when evaluated on MSCE yields very similar conclusions to our evaluation of NN. We show the Pareto optimality Tables (Table \ref{tbl: XGBsummaryMSCE}, Table \ref{tbl: NNsummaryMSCE}) as well as the aggregate results over the ACS datasets and the aggregate results in Figure  \ref{fig:MSCEFrontiers}. Note that because we tuned some of the methods for ECCE and are evaluating them for MSCE and KCE in the respective sections, the comparisons between the calibration error metrics are not entirely sensible but we show them anyways for comprehensiveness. 

Again, when we look across all datasets, it is clear that when in the "None" data regime, there is little room for improving ECCE beyond the baseline of tuning for accuracy. When looking at the ACS datasets, we observe that the group robust training methods are consistently Pareto optimal across both XGB and NN implementations, which corroborates our second conclusion of Section \ref{Section: Framework}. Lastly, the conclusion of model inductive bias is clear from here from the accuracy view, but less so from the frontier perspective. We attribute this discrepancy to the nuances of optimizing for ECCE and evaluating by MSCE. 

\begin{center}
\begin{table}[h!]
\begin{tabulary}{\linewidth}{||LCCL||} 
 \hline
 Data Availability & \# Methods Tested & \# Times Optimal Across Datasets & Best Pareto Optimal Method \\ [0.5ex] 
 \hline\hline
 Train+Val+Inf & 2 & 7 & Use group (5/7) \\ 
 \hline
 Train+Val & 1 & 3 & Group-robust calibration training (3/3) \\
 \hline
 Val & 3 & 7 & Group-robust calibrator (4/7) \\
 \hline
 None & 10 & 10 & Tune for accuracy (3/10) \\ [1ex] 
 \hline
\end{tabulary}
\caption{Summarizing benchmarks across 10 datasets for XGB (Using MSCE)}
 \label{tbl: XGBsummaryMSCE}
\end{table}
\end{center}

\begin{center}
\begin{table}[h!]
\begin{tabulary}{\linewidth}{||LCCL||} 
 \hline
 Data Availability & \# Methods Tested & \# Times Optimal Across Datasets & Best Pareto Optimal Method \\ [0.5ex] 
 \hline\hline
 Train+Val+Inf & 2 & 8 & Use group (4/8) \\ 
 \hline
 Train+Val & 2 & 7 & Group-robust calibration training (4/7) \\
 \hline
 Val & 3 & 1 & Group-robust calibrator (1/1) \\
 \hline
 None & 12 & 10 & Multicalibration (4/10) \\ [1ex] 
 \hline
\end{tabulary}
\caption{Summarizing benchmarks across 10 datasets for NN (Using MSCE)}
\label{tbl: NNsummaryMSCE}
\end{table}
\end{center}

\subsection{Pareto Frontier Tables for KCE} \label{App: Results for KCE}

\begin{center}
    \begin{figure}[h!] 
    \caption{Pareto frontiers when aggregating data across ACS data (top) and All data (bottom). The degree of predictive parity fairness is plotted on the Y-axis and data requirement on the X-axis. Each point represents a unique fairness intervention method for predictive parity (points close to the bottom left-hand corner are better). We have labeled the Pareto optimal points as well as the baseline points "Tune for Acc." Model performance is considered by observing the accuracy range. We have also added the "Other" category for QPD, but do not consider it to be part of the Pareto frontier.}
    \label{fig:KCEFrontiers}
    \includegraphics[width=\textwidth]{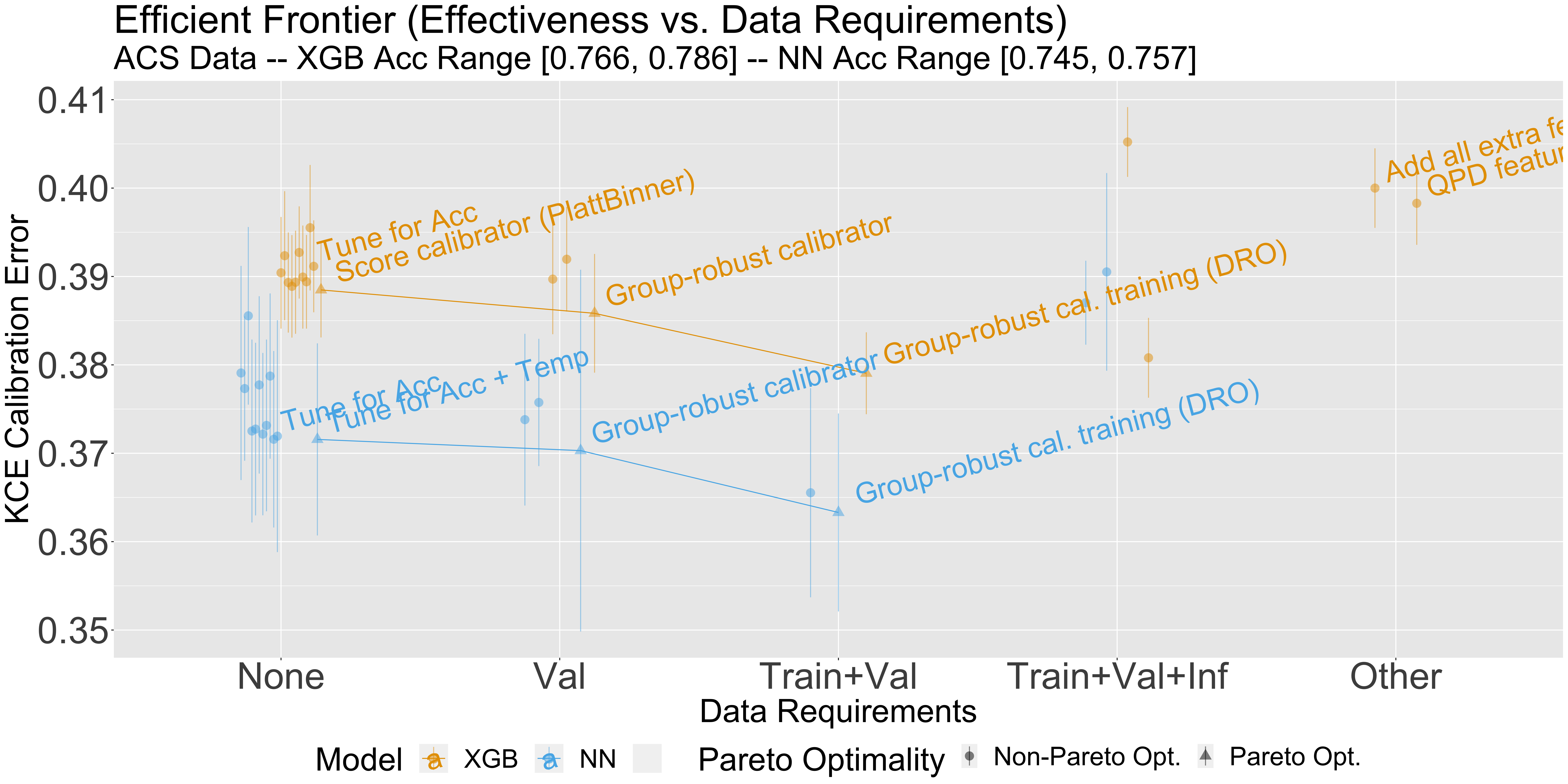}
    \includegraphics[width=\textwidth]{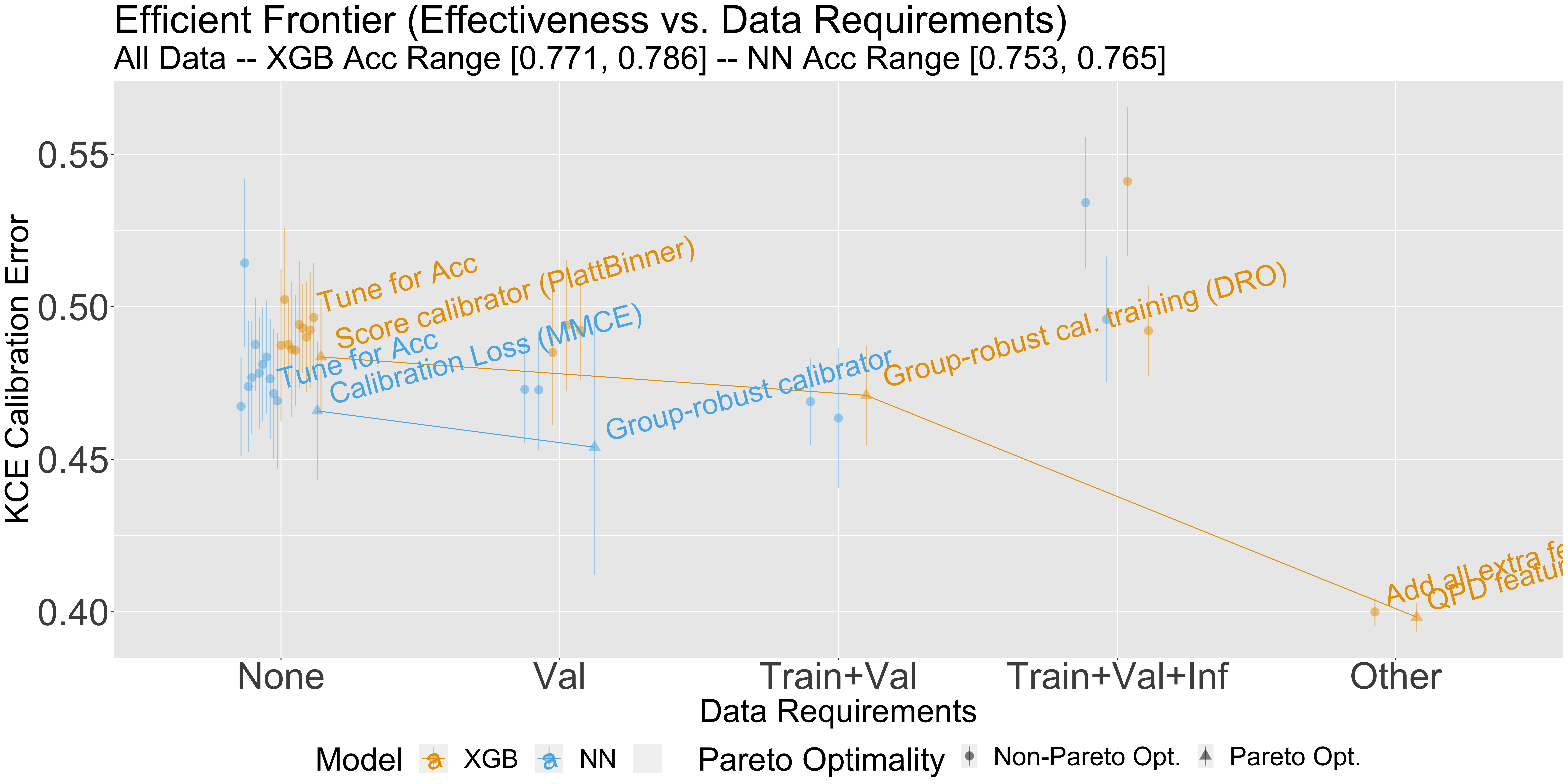}
    \end{figure}
\end{center}

In this section, we show the same thing as in Appendix \ref{App: Results for MSCE}. Again, we note that because we tuned some of the methods for ECCE and are evaluating them for KCE, the comparisons between the calibration error metrics are not entirely sensible but we show them anyways for comprehensiveness. In \ref{App: CE Metrics} we also remarked that we consider KCE to be the least reliable calibration error metric, as although it has intriguing theoretical properties \cite{BlasiokCalibration} practically evaluating this metric in a reasonable amount of time calls for random sampling, which we find to be an undesirable property in general and particularly undesirable when making sensitive measurements in real applications such as fairness. We show the Pareto frontier tables in Table \ref{tbl: XGBsummaryKCE} and Table \ref{tbl: NNsummaryKCE} as well as the aggregate results over the ACS datasets and the aggregate results in Figure \ref{fig:KCEFrontiers}.

For KCE as well, it is clear that in the "None" data regime, it is difficult to improve fairness beyond the baseline of tuning for accuracy. We again see in the ACS and all dataset aggregations that group robust training is Pareto optimal and thus is a promising path forward to predictive parity fairness without inference. In the "All" datatset aggregation, we do not see that it works well for neural networks, which we attribute to the small dataset sizes as shown in Table \ref{tbl: datasets} and the fact that neural networks generally do not work as well with fewer datapoints. Lastly, we interestingly see that the model inductive bias conclusions is reversed in this setting of KCE measurement as opposed to ECCE measurement. Again, we do not consider this to be the most reliable calibration error measurement due to the randomness of the sampling. Nonetheless, this represents an intriguing demonstration that specifying the right calibration error metric is critical to making conclusions and picking the right intervention model.

\begin{center}
\begin{table}[h!]
\begin{tabulary}{\linewidth}{||LCCL||} 
 \hline
 Data Availability & \# Methods Tested & \# Times Optimal Across Datasets & Best Pareto Optimal Method \\ [0.5ex] 
 \hline\hline
 Train+Val+Inf & 2 & 7 & Use group (5/7) \\ 
 \hline
 Train+Val & 1 & 3 & Group-robust calibration training (3/3) \\
 \hline
 Val & 3 & 7 & Group-robust calibrator (4/7) \\
 \hline
 None & 10 & 10 & Tune for accuracy (3/10) \\ [1ex] 
 \hline
\end{tabulary}
\caption{Summarizing benchmarks across 10 datasets for XGB (Using KCE)}
\label{tbl: XGBsummaryKCE}
\end{table}
\end{center}

\begin{center}
\begin{table}[h!]
\begin{tabulary}{\linewidth}{||LCCL||} 
 \hline
 Data Availability & \# Methods Tested & \# Times Optimal Across Datasets & Best Pareto Optimal Method \\ [0.5ex] 
 \hline\hline
 Train+Val+Inf & 2 & 8 & Use group (4/8) \\ 
 \hline
 Train+Val & 2 & 7 & Group-robust calibration training (4/7) \\
 \hline
 Val & 3 & 1 & Group-robust calibrator (1/1) \\
 \hline
 None & 12 & 10 & Multicalibration (4/10) \\ [1ex] 
 \hline
\end{tabulary}
 
\caption{Summarizing benchmarks across 10 datasets for NN (Using KCE)}
\label{tbl: NNsummaryKCE}
\end{table}
\end{center}

%% file: main.bbl

\begin{thebibliography}{73}


\ifx \showCODEN    \undefined \def \showCODEN     #1{\unskip}     \fi
\ifx \showDOI      \undefined \def \showDOI       #1{#1}\fi
\ifx \showISBNx    \undefined \def \showISBNx     #1{\unskip}     \fi
\ifx \showISBNxiii \undefined \def \showISBNxiii  #1{\unskip}     \fi
\ifx \showISSN     \undefined \def \showISSN      #1{\unskip}     \fi
\ifx \showLCCN     \undefined \def \showLCCN      #1{\unskip}     \fi
\ifx \shownote     \undefined \def \shownote      #1{#1}          \fi
\ifx \showarticletitle \undefined \def \showarticletitle #1{#1}   \fi
\ifx \showURL      \undefined \def \showURL       {\relax}        \fi
\providecommand\bibfield[2]{#2}
\providecommand\bibinfo[2]{#2}
\providecommand\natexlab[1]{#1}
\providecommand\showeprint[2][]{arXiv:#2}

\bibitem[Andrus et~al\mbox{.}(2021)]%
        {AndrusSpitzer}
\bibfield{author}{\bibinfo{person}{McKane Andrus}, \bibinfo{person}{Elena
  Spitzer}, \bibinfo{person}{Jeffrey Brown}, {and} \bibinfo{person}{Alice
  Xiang}.} \bibinfo{year}{2021}\natexlab{}.
\newblock \showarticletitle{What We Can't Measure, We Can't Understand:
  Challenges to Demographic Data Procurement in the Pursuit of Fairness}. In
  \bibinfo{booktitle}{\emph{Proceedings of the 2021 ACM Conference on Fairness,
  Accountability, and Transparency}} (Virtual Event, Canada)
  \emph{(\bibinfo{series}{FAccT '21})}. \bibinfo{publisher}{Association for
  Computing Machinery}, \bibinfo{address}{New York, NY, USA},
  \bibinfo{pages}{249–260}.
\newblock
\showISBNx{9781450383097}
\urldef\tempurl%
\url{https://doi.org/10.1145/3442188.3445888}
\showDOI{\tempurl}


\bibitem[Andrus and Villeneuve(2022)]%
        {AndrusDemographic}
\bibfield{author}{\bibinfo{person}{McKane Andrus} {and} \bibinfo{person}{Sarah
  Villeneuve}.} \bibinfo{year}{2022}\natexlab{}.
\newblock \showarticletitle{Demographic-Reliant Algorithmic Fairness:
  Characterizing the Risks of Demographic Data Collection in the Pursuit of
  Fairness}. In \bibinfo{booktitle}{\emph{2022 ACM Conference on Fairness,
  Accountability, and Transparency}} (Seoul, Republic of Korea)
  \emph{(\bibinfo{series}{FAccT '22})}. \bibinfo{publisher}{Association for
  Computing Machinery}, \bibinfo{address}{New York, NY, USA},
  \bibinfo{pages}{1709–1721}.
\newblock
\showISBNx{9781450393522}
\urldef\tempurl%
\url{https://doi.org/10.1145/3531146.3533226}
\showDOI{\tempurl}


\bibitem[Angwin et~al\mbox{.}(2016)]%
        {COMPAS_angwin_larson_mattu_kirchner_2016}
\bibfield{author}{\bibinfo{person}{Julia Angwin}, \bibinfo{person}{Jeff
  Larson}, \bibinfo{person}{Surya Mattu}, {and} \bibinfo{person}{Lauren
  Kirchner}.} \bibinfo{year}{2016}\natexlab{}.
\newblock \bibinfo{title}{Machine bias}.
\newblock
\newblock
\urldef\tempurl%
\url{https://www.propublica.org/article/machine-bias-risk-assessments-in-criminal-sentencing}
\showURL{%
\tempurl}


\bibitem[Arrieta-Ibarra et~al\mbox{.}(2022)]%
        {arrieta2022metrics}
\bibfield{author}{\bibinfo{person}{Imanol Arrieta-Ibarra},
  \bibinfo{person}{Paman Gujral}, \bibinfo{person}{Jonathan Tannen},
  \bibinfo{person}{Mark Tygert}, {and} \bibinfo{person}{Cherie Xu}.}
  \bibinfo{year}{2022}\natexlab{}.
\newblock \showarticletitle{Metrics of calibration for probabilistic
  predictions}.
\newblock \bibinfo{journal}{\emph{arXiv preprint arXiv:2205.09680}}
  (\bibinfo{year}{2022}).
\newblock


\bibitem[Bakalar et~al\mbox{.}(2021)]%
        {bakalarFairnessOnTheGround}
\bibfield{author}{\bibinfo{person}{Chloé Bakalar}, \bibinfo{person}{Renata
  Barreto}, \bibinfo{person}{Stevie Bergman}, \bibinfo{person}{Miranda Bogen},
  \bibinfo{person}{Bobbie Chern}, \bibinfo{person}{Sam Corbett-Davies},
  \bibinfo{person}{Melissa Hall}, \bibinfo{person}{Isabel Kloumann},
  \bibinfo{person}{Michelle Lam}, \bibinfo{person}{Joaquin~Quiñonero Candela},
  \bibinfo{person}{Manish Raghavan}, \bibinfo{person}{Joshua Simons},
  \bibinfo{person}{Jonathan Tannen}, \bibinfo{person}{Edmund Tong},
  \bibinfo{person}{Kate Vredenburgh}, {and} \bibinfo{person}{Jiejing Zhao}.}
  \bibinfo{year}{2021}\natexlab{}.
\newblock \bibinfo{title}{Fairness On The Ground: Applying Algorithmic Fairness
  Approaches to Production Systems}.
\newblock
\newblock
\urldef\tempurl%
\url{https://doi.org/10.48550/ARXIV.2103.06172}
\showDOI{\tempurl}


\bibitem[Barocas et~al\mbox{.}(2019)]%
        {barocas-hardt-narayanan}
\bibfield{author}{\bibinfo{person}{Solon Barocas}, \bibinfo{person}{Moritz
  Hardt}, {and} \bibinfo{person}{Arvind Narayanan}.}
  \bibinfo{year}{2019}\natexlab{}.
\newblock \bibinfo{booktitle}{\emph{Fairness and Machine Learning}}.
\newblock \bibinfo{publisher}{fairmlbook.org}.
\newblock
\newblock
\shownote{\url{http://www.fairmlbook.org}}.


\bibitem[Bent(2019)]%
        {bent2019algorithmic}
\bibfield{author}{\bibinfo{person}{Jason~R Bent}.}
  \bibinfo{year}{2019}\natexlab{}.
\newblock \showarticletitle{Is algorithmic affirmative action legal}.
\newblock \bibinfo{journal}{\emph{Geo. LJ}}  \bibinfo{volume}{108}
  (\bibinfo{year}{2019}), \bibinfo{pages}{803}.
\newblock


\bibitem[Bogen et~al\mbox{.}(2020)]%
        {BogenAwareness}
\bibfield{author}{\bibinfo{person}{Miranda Bogen}, \bibinfo{person}{Aaron
  Rieke}, {and} \bibinfo{person}{Shazeda Ahmed}.}
  \bibinfo{year}{2020}\natexlab{}.
\newblock \showarticletitle{Awareness in Practice: Tensions in Access to
  Sensitive Attribute Data for Antidiscrimination}. In
  \bibinfo{booktitle}{\emph{Proceedings of the 2020 Conference on Fairness,
  Accountability, and Transparency}} (Barcelona, Spain)
  \emph{(\bibinfo{series}{FAT* '20})}. \bibinfo{publisher}{Association for
  Computing Machinery}, \bibinfo{address}{New York, NY, USA},
  \bibinfo{pages}{492–500}.
\newblock
\showISBNx{9781450369367}
\urldef\tempurl%
\url{https://doi.org/10.1145/3351095.3372877}
\showDOI{\tempurl}


\bibitem[Błasiok et~al\mbox{.}(2022)]%
        {BlasiokCalibration}
\bibfield{author}{\bibinfo{person}{Jarosław Błasiok},
  \bibinfo{person}{Parikshit Gopalan}, \bibinfo{person}{Lunjia Hu}, {and}
  \bibinfo{person}{Preetum Nakkiran}.} \bibinfo{year}{2022}\natexlab{}.
\newblock \bibinfo{title}{A Unifying Theory of Distance from Calibration}.
\newblock
\newblock
\urldef\tempurl%
\url{https://doi.org/10.48550/ARXIV.2211.16886}
\showDOI{\tempurl}


\bibitem[Cai et~al\mbox{.}(2022)]%
        {caiAdaptiveSampling}
\bibfield{author}{\bibinfo{person}{William Cai}, \bibinfo{person}{Ro
  Encarnacion}, \bibinfo{person}{Bobbie Chern}, \bibinfo{person}{Sam
  Corbett-Davies}, \bibinfo{person}{Miranda Bogen}, \bibinfo{person}{Stevie
  Bergman}, {and} \bibinfo{person}{Sharad Goel}.}
  \bibinfo{year}{2022}\natexlab{}.
\newblock \showarticletitle{Adaptive Sampling Strategies to Construct Equitable
  Training Datasets}. In \bibinfo{booktitle}{\emph{2022 ACM Conference on
  Fairness, Accountability, and Transparency}} (Seoul, Republic of Korea)
  \emph{(\bibinfo{series}{FAccT '22})}. \bibinfo{publisher}{Association for
  Computing Machinery}, \bibinfo{address}{New York, NY, USA},
  \bibinfo{pages}{1467–1478}.
\newblock
\showISBNx{9781450393522}
\urldef\tempurl%
\url{https://doi.org/10.1145/3531146.3533203}
\showDOI{\tempurl}


\bibitem[Cai et~al\mbox{.}(2020)]%
        {CaiFairAllocation}
\bibfield{author}{\bibinfo{person}{William Cai}, \bibinfo{person}{Johann
  Gaebler}, \bibinfo{person}{Nikhil Garg}, {and} \bibinfo{person}{Sharad
  Goel}.} \bibinfo{year}{2020}\natexlab{}.
\newblock \showarticletitle{Fair Allocation through Selective Information
  Acquisition}. In \bibinfo{booktitle}{\emph{Proceedings of the AAAI/ACM
  Conference on AI, Ethics, and Society}} (New York, NY, USA)
  \emph{(\bibinfo{series}{AIES '20})}. \bibinfo{publisher}{Association for
  Computing Machinery}, \bibinfo{address}{New York, NY, USA},
  \bibinfo{pages}{22–28}.
\newblock
\showISBNx{9781450371100}
\urldef\tempurl%
\url{https://doi.org/10.1145/3375627.3375823}
\showDOI{\tempurl}


\bibitem[Celis et~al\mbox{.}(2019)]%
        {Celis2019}
\bibfield{author}{\bibinfo{person}{L.~Elisa Celis}, \bibinfo{person}{Lingxiao
  Huang}, \bibinfo{person}{Vijay Keswani}, {and} \bibinfo{person}{Nisheeth~K.
  Vishnoi}.} \bibinfo{year}{2019}\natexlab{}.
\newblock \showarticletitle{Classification with Fairness Constraints: A
  Meta-Algorithm with Provable Guarantees} \emph{(\bibinfo{series}{FAT* '19})}.
  \bibinfo{publisher}{Association for Computing Machinery},
  \bibinfo{address}{New York, NY, USA}, \bibinfo{pages}{319–328}.
\newblock
\showISBNx{9781450361255}
\urldef\tempurl%
\url{https://doi.org/10.1145/3287560.3287586}
\showDOI{\tempurl}


\bibitem[Chen and Guestrin(2016)]%
        {Chen:2016:XST:2939672.2939785}
\bibfield{author}{\bibinfo{person}{Tianqi Chen} {and} \bibinfo{person}{Carlos
  Guestrin}.} \bibinfo{year}{2016}\natexlab{}.
\newblock \showarticletitle{{XGBoost}: A Scalable Tree Boosting System}. In
  \bibinfo{booktitle}{\emph{Proceedings of the 22nd ACM SIGKDD International
  Conference on Knowledge Discovery and Data Mining}} (San Francisco,
  California, USA) \emph{(\bibinfo{series}{KDD '16})}.
  \bibinfo{publisher}{ACM}, \bibinfo{address}{New York, NY, USA},
  \bibinfo{pages}{785--794}.
\newblock
\showISBNx{978-1-4503-4232-2}
\urldef\tempurl%
\url{https://doi.org/10.1145/2939672.2939785}
\showDOI{\tempurl}


\bibitem[Chen et~al\mbox{.}(2022)]%
        {ChenTestingSurvey}
\bibfield{author}{\bibinfo{person}{Zhenpeng Chen}, \bibinfo{person}{Jie~M.
  Zhang}, \bibinfo{person}{Max Hort}, \bibinfo{person}{Federica Sarro}, {and}
  \bibinfo{person}{Mark Harman}.} \bibinfo{year}{2022}\natexlab{}.
\newblock \bibinfo{title}{Fairness Testing: A Comprehensive Survey and Analysis
  of Trends}.
\newblock
\newblock
\urldef\tempurl%
\url{https://doi.org/10.48550/ARXIV.2207.10223}
\showDOI{\tempurl}


\bibitem[Chouldechova(2017)]%
        {Chouldechova17}
\bibfield{author}{\bibinfo{person}{Alexandra Chouldechova}.}
  \bibinfo{year}{2017}\natexlab{}.
\newblock \showarticletitle{Fair Prediction with Disparate Impact: {A} Study of
  Bias in Recidivism Prediction Instruments}.
\newblock \bibinfo{journal}{\emph{Big Data}} \bibinfo{volume}{5},
  \bibinfo{number}{2} (\bibinfo{year}{2017}), \bibinfo{pages}{153--163}.
\newblock
\urldef\tempurl%
\url{https://doi.org/10.1089/big.2016.0047}
\showDOI{\tempurl}


\bibitem[Colmenarejo et~al\mbox{.}(2022)]%
        {Bringas_Colmenarejo_2022}
\bibfield{author}{\bibinfo{person}{Alejandra~Bringas Colmenarejo},
  \bibinfo{person}{Luca Nannini}, \bibinfo{person}{Alisa Rieger},
  \bibinfo{person}{Kristen~M. Scott}, \bibinfo{person}{Xuan Zhao},
  \bibinfo{person}{Gourab~K Patro}, \bibinfo{person}{Gjergji Kasneci}, {and}
  \bibinfo{person}{Katharina Kinder-Kurlanda}.}
  \bibinfo{year}{2022}\natexlab{}.
\newblock \showarticletitle{Fairness in Agreement With European Values}. In
  \bibinfo{booktitle}{\emph{Proceedings of the 2022 {AAAI}/{ACM} Conference on
  {AI}, Ethics, and Society}}. \bibinfo{publisher}{{ACM}}.
\newblock
\urldef\tempurl%
\url{https://doi.org/10.1145/3514094.3534158}
\showDOI{\tempurl}


\bibitem[Cotter et~al\mbox{.}(2019)]%
        {cotterNonconvexConstrained}
\bibfield{author}{\bibinfo{person}{Andrew Cotter}, \bibinfo{person}{Heinrich
  Jiang}, {and} \bibinfo{person}{Karthik Sridharan}.}
  \bibinfo{year}{2019}\natexlab{}.
\newblock \showarticletitle{Two-Player Games for Efficient Non-Convex
  Constrained Optimization}. In \bibinfo{booktitle}{\emph{Proceedings of the
  30th International Conference on Algorithmic Learning Theory}}
  \emph{(\bibinfo{series}{Proceedings of Machine Learning Research},
  Vol.~\bibinfo{volume}{98})}, \bibfield{editor}{\bibinfo{person}{Aurélien
  Garivier} {and} \bibinfo{person}{Satyen Kale}} (Eds.).
  \bibinfo{publisher}{PMLR}, \bibinfo{pages}{300--332}.
\newblock
\urldef\tempurl%
\url{https://proceedings.mlr.press/v98/cotter19a.html}
\showURL{%
\tempurl}


\bibitem[Croak and Gennai(2023)]%
        {croak_gennai_2023}
\bibfield{author}{\bibinfo{person}{Marian Croak} {and} \bibinfo{person}{Jen
  Gennai}.} \bibinfo{year}{2023}\natexlab{}.
\newblock \bibinfo{title}{Responsible ai: Looking back at 2022, and to the
  future}.
\newblock
\newblock
\urldef\tempurl%
\url{https://blog.google/technology/ai/responsible-ai-looking-back-at-2022-and-to-the-future/}
\showURL{%
\tempurl}


\bibitem[DiCiccio et~al\mbox{.}(2022)]%
        {diciccio2022predictive}
\bibfield{author}{\bibinfo{person}{Cyrus DiCiccio}, \bibinfo{person}{Brian
  Hsu}, \bibinfo{person}{YinYin Yu}, \bibinfo{person}{Preetam Nandy}, {and}
  \bibinfo{person}{Kinjal Basu}.} \bibinfo{year}{2022}\natexlab{}.
\newblock \showarticletitle{Predictive Rate Parity Testing and Mitigation}.
\newblock \bibinfo{journal}{\emph{arXiv preprint arXiv:2204.05947}}
  (\bibinfo{year}{2022}).
\newblock


\bibitem[Dieterich et~al\mbox{.}(2016)]%
        {dieterich2016compas}
\bibfield{author}{\bibinfo{person}{William Dieterich},
  \bibinfo{person}{Christina Mendoza}, {and} \bibinfo{person}{Tim Brennan}.}
  \bibinfo{year}{2016}\natexlab{}.
\newblock \showarticletitle{COMPAS risk scales: Demonstrating accuracy equity
  and predictive parity}.
\newblock \bibinfo{journal}{\emph{Northpointe Inc}} \bibinfo{volume}{7},
  \bibinfo{number}{7.4} (\bibinfo{year}{2016}), \bibinfo{pages}{1}.
\newblock


\bibitem[Ding et~al\mbox{.}(2021)]%
        {DingRetiring}
\bibfield{author}{\bibinfo{person}{Frances Ding}, \bibinfo{person}{Moritz
  Hardt}, \bibinfo{person}{John Miller}, {and} \bibinfo{person}{Ludwig
  Schmidt}.} \bibinfo{year}{2021}\natexlab{}.
\newblock \showarticletitle{Retiring Adult: New Datasets for Fair Machine
  Learning}. In \bibinfo{booktitle}{\emph{Advances in Neural Information
  Processing Systems}}, \bibfield{editor}{\bibinfo{person}{M.~Ranzato},
  \bibinfo{person}{A.~Beygelzimer}, \bibinfo{person}{Y.~Dauphin},
  \bibinfo{person}{P.S. Liang}, {and} \bibinfo{person}{J.~Wortman Vaughan}}
  (Eds.), Vol.~\bibinfo{volume}{34}. \bibinfo{publisher}{Curran Associates,
  Inc.}, \bibinfo{pages}{6478--6490}.
\newblock
\urldef\tempurl%
\url{https://proceedings.neurips.cc/paper/2021/file/32e54441e6382a7fbacbbbaf3c450059-Paper.pdf}
\showURL{%
\tempurl}


\bibitem[Dua and Graff(2017)]%
        {DuaUCI}
\bibfield{author}{\bibinfo{person}{Dheeru Dua} {and} \bibinfo{person}{Casey
  Graff}.} \bibinfo{year}{2017}\natexlab{}.
\newblock \bibinfo{title}{{UCI} Machine Learning Repository}.
\newblock
\newblock
\urldef\tempurl%
\url{http://archive.ics.uci.edu/ml}
\showURL{%
\tempurl}


\bibitem[Durfee et~al\mbox{.}(2022)]%
        {DurfeeHetCal}
\bibfield{author}{\bibinfo{person}{David Durfee}, \bibinfo{person}{Aman Gupta},
  {and} \bibinfo{person}{Kinjal Basu}.} \bibinfo{year}{2022}\natexlab{}.
\newblock \bibinfo{title}{Heterogeneous Calibration: A post-hoc model-agnostic
  framework for improved generalization}.
\newblock
\newblock
\urldef\tempurl%
\url{https://doi.org/10.48550/ARXIV.2202.04837}
\showDOI{\tempurl}


\bibitem[Gardner et~al\mbox{.}(2022)]%
        {GardnerXGB}
\bibfield{author}{\bibinfo{person}{Josh Gardner}, \bibinfo{person}{Zoran
  Popović}, {and} \bibinfo{person}{Ludwig Schmidt}.}
  \bibinfo{year}{2022}\natexlab{}.
\newblock \bibinfo{title}{Subgroup Robustness Grows On Trees: An Empirical
  Baseline Investigation}.
\newblock
\newblock
\urldef\tempurl%
\url{https://doi.org/10.48550/ARXIV.2211.12703}
\showDOI{\tempurl}


\bibitem[Ghosh et~al\mbox{.}(2022)]%
        {Ghosh_2022}
\bibfield{author}{\bibinfo{person}{Avijit Ghosh}, \bibinfo{person}{Aalok
  Shanbhag}, {and} \bibinfo{person}{Christo Wilson}.}
  \bibinfo{year}{2022}\natexlab{}.
\newblock \showarticletitle{{FairCanary}}. In
  \bibinfo{booktitle}{\emph{Proceedings of the 2022 {AAAI}/{ACM} Conference on
  {AI}, Ethics, and Society}}. \bibinfo{publisher}{{ACM}}.
\newblock
\urldef\tempurl%
\url{https://doi.org/10.1145/3514094.3534157}
\showDOI{\tempurl}


\bibitem[Grgi{\'c}-Hla{\v{c}}a et~al\mbox{.}(2018)]%
        {grgic2018beyond}
\bibfield{author}{\bibinfo{person}{Nina Grgi{\'c}-Hla{\v{c}}a},
  \bibinfo{person}{Muhammad~Bilal Zafar}, \bibinfo{person}{Krishna~P Gummadi},
  {and} \bibinfo{person}{Adrian Weller}.} \bibinfo{year}{2018}\natexlab{}.
\newblock \showarticletitle{Beyond distributive fairness in algorithmic
  decision making: Feature selection for procedurally fair learning}. In
  \bibinfo{booktitle}{\emph{Proceedings of the AAAI Conference on Artificial
  Intelligence}}, Vol.~\bibinfo{volume}{32}.
\newblock


\bibitem[Gruber and Buettner(2022)]%
        {Gruber2022}
\bibfield{author}{\bibinfo{person}{Sebastian Gruber} {and}
  \bibinfo{person}{Florian Buettner}.} \bibinfo{year}{2022}\natexlab{}.
\newblock \showarticletitle{Better Uncertainty Calibration via Proper Scores
  for Classification and Beyond}. In \bibinfo{booktitle}{\emph{Advances in
  Neural Information Processing Systems 35: Annual Conference on Neural
  Information Processing Systems 2022, NeurIPS 2022}}.
\newblock


\bibitem[Guo et~al\mbox{.}(2017)]%
        {guoCalibration}
\bibfield{author}{\bibinfo{person}{Chuan Guo}, \bibinfo{person}{Geoff Pleiss},
  \bibinfo{person}{Yu Sun}, {and} \bibinfo{person}{Kilian~Q. Weinberger}.}
  \bibinfo{year}{2017}\natexlab{}.
\newblock \showarticletitle{On Calibration of Modern Neural Networks}. In
  \bibinfo{booktitle}{\emph{Proceedings of the 34th International Conference on
  Machine Learning}} \emph{(\bibinfo{series}{Proceedings of Machine Learning
  Research}, Vol.~\bibinfo{volume}{70})},
  \bibfield{editor}{\bibinfo{person}{Doina Precup} {and}
  \bibinfo{person}{Yee~Whye Teh}} (Eds.). \bibinfo{publisher}{PMLR},
  \bibinfo{pages}{1321--1330}.
\newblock
\urldef\tempurl%
\url{https://proceedings.mlr.press/v70/guo17a.html}
\showURL{%
\tempurl}


\bibitem[Hardt et~al\mbox{.}(2016)]%
        {Hardtetal_NIPS2016}
\bibfield{author}{\bibinfo{person}{Moritz Hardt}, \bibinfo{person}{Eric Price},
  \bibinfo{person}{Eric Price}, {and} \bibinfo{person}{Nati Srebro}.}
  \bibinfo{year}{2016}\natexlab{}.
\newblock \showarticletitle{Equality of Opportunity in Supervised Learning}.
\newblock In \bibinfo{booktitle}{\emph{Advances in Neural Information
  Processing Systems 29}}, \bibfield{editor}{\bibinfo{person}{D.~D. Lee},
  \bibinfo{person}{M.~Sugiyama}, \bibinfo{person}{U.~V. Luxburg},
  \bibinfo{person}{I.~Guyon}, {and} \bibinfo{person}{R.~Garnett}} (Eds.).
  \bibinfo{publisher}{Curran Associates, Inc.}, \bibinfo{pages}{3315--3323}.
\newblock


\bibitem[Hashimoto et~al\mbox{.}(2018)]%
        {pmlr-v80-hashimoto18a}
\bibfield{author}{\bibinfo{person}{Tatsunori Hashimoto}, \bibinfo{person}{Megha
  Srivastava}, \bibinfo{person}{Hongseok Namkoong}, {and}
  \bibinfo{person}{Percy Liang}.} \bibinfo{year}{2018}\natexlab{}.
\newblock \showarticletitle{Fairness Without Demographics in Repeated Loss
  Minimization}. In \bibinfo{booktitle}{\emph{Proceedings of the 35th
  International Conference on Machine Learning}}
  \emph{(\bibinfo{series}{Proceedings of Machine Learning Research},
  Vol.~\bibinfo{volume}{80})}, \bibfield{editor}{\bibinfo{person}{Jennifer Dy}
  {and} \bibinfo{person}{Andreas Krause}} (Eds.). \bibinfo{publisher}{PMLR},
  \bibinfo{pages}{1929--1938}.
\newblock
\urldef\tempurl%
\url{https://proceedings.mlr.press/v80/hashimoto18a.html}
\showURL{%
\tempurl}


\bibitem[Hebert-Johnson et~al\mbox{.}(2018)]%
        {pmlr-v80-hebert-johnson18a}
\bibfield{author}{\bibinfo{person}{Ursula Hebert-Johnson},
  \bibinfo{person}{Michael Kim}, \bibinfo{person}{Omer Reingold}, {and}
  \bibinfo{person}{Guy Rothblum}.} \bibinfo{year}{2018}\natexlab{}.
\newblock \showarticletitle{Multicalibration: Calibration for the
  ({C}omputationally-Identifiable) Masses}. In
  \bibinfo{booktitle}{\emph{Proceedings of the 35th International Conference on
  Machine Learning}} \emph{(\bibinfo{series}{Proceedings of Machine Learning
  Research}, Vol.~\bibinfo{volume}{80})},
  \bibfield{editor}{\bibinfo{person}{Jennifer Dy} {and}
  \bibinfo{person}{Andreas Krause}} (Eds.). \bibinfo{publisher}{PMLR},
  \bibinfo{pages}{1939--1948}.
\newblock
\urldef\tempurl%
\url{https://proceedings.mlr.press/v80/hebert-johnson18a.html}
\showURL{%
\tempurl}


\bibitem[Hort et~al\mbox{.}(2022)]%
        {HortMitigationSurvey}
\bibfield{author}{\bibinfo{person}{Max Hort}, \bibinfo{person}{Zhenpeng Chen},
  \bibinfo{person}{Jie~M. Zhang}, \bibinfo{person}{Federica Sarro}, {and}
  \bibinfo{person}{Mark Harman}.} \bibinfo{year}{2022}\natexlab{}.
\newblock \bibinfo{title}{Bias Mitigation for Machine Learning Classifiers: A
  Comprehensive Survey}.
\newblock
\newblock
\urldef\tempurl%
\url{https://doi.org/10.48550/ARXIV.2207.07068}
\showDOI{\tempurl}


\bibitem[Hsu et~al\mbox{.}(2022)]%
        {Hsu2022}
\bibfield{author}{\bibinfo{person}{Brian Hsu}, \bibinfo{person}{Rahul
  Mazumder}, \bibinfo{person}{Preetam Nandy}, {and} \bibinfo{person}{Kinjal
  Basu}.} \bibinfo{year}{2022}\natexlab{}.
\newblock \showarticletitle{Pushing the limits of fairness impossibility: Who's
  the fairest of them all?}. In \bibinfo{booktitle}{\emph{Advances in Neural
  Information Processing Systems 35: Annual Conference on Neural Information
  Processing Systems 2022, NeurIPS 2022}}.
\newblock


\bibitem[Karandikar et~al\mbox{.}(2021)]%
        {KarandikarSoftmax}
\bibfield{author}{\bibinfo{person}{Archit Karandikar},
  \bibinfo{person}{Nicholas Cain}, \bibinfo{person}{Dustin Tran},
  \bibinfo{person}{Balaji Lakshminarayanan}, \bibinfo{person}{Jonathon Shlens},
  \bibinfo{person}{Michael~C Mozer}, {and} \bibinfo{person}{Becca Roelofs}.}
  \bibinfo{year}{2021}\natexlab{}.
\newblock \showarticletitle{Soft Calibration Objectives for Neural Networks}.
  In \bibinfo{booktitle}{\emph{Advances in Neural Information Processing
  Systems}}, \bibfield{editor}{\bibinfo{person}{M.~Ranzato},
  \bibinfo{person}{A.~Beygelzimer}, \bibinfo{person}{Y.~Dauphin},
  \bibinfo{person}{P.S. Liang}, {and} \bibinfo{person}{J.~Wortman Vaughan}}
  (Eds.), Vol.~\bibinfo{volume}{34}. \bibinfo{publisher}{Curran Associates,
  Inc.}, \bibinfo{pages}{29768--29779}.
\newblock
\urldef\tempurl%
\url{https://proceedings.neurips.cc/paper/2021/file/f8905bd3df64ace64a68e154ba72f24c-Paper.pdf}
\showURL{%
\tempurl}


\bibitem[Kim et~al\mbox{.}(2018)]%
        {DBLP:journals/corr/abs-1805-12317}
\bibfield{author}{\bibinfo{person}{Michael~P. Kim}, \bibinfo{person}{Amirata
  Ghorbani}, {and} \bibinfo{person}{James~Y. Zou}.}
  \bibinfo{year}{2018}\natexlab{}.
\newblock \showarticletitle{Multiaccuracy: Black-Box Post-Processing for
  Fairness in Classification}.
\newblock \bibinfo{journal}{\emph{CoRR}}  \bibinfo{volume}{abs/1805.12317}
  (\bibinfo{year}{2018}).
\newblock
\showeprint[arXiv]{1805.12317}
\urldef\tempurl%
\url{http://arxiv.org/abs/1805.12317}
\showURL{%
\tempurl}


\bibitem[Kloumann and Tannen(2021)]%
        {kloumann_tannen_2021}
\bibfield{author}{\bibinfo{person}{Isabel Kloumann} {and}
  \bibinfo{person}{Jonathan Tannen}.} \bibinfo{year}{2021}\natexlab{}.
\newblock \bibinfo{title}{Building AI that works better for everyone}.
\newblock
\newblock
\urldef\tempurl%
\url{https://about.fb.com/news/2021/03/building-ai-that-works-better-for-everyone/}
\showURL{%
\tempurl}


\bibitem[Kramer(2021)]%
        {kramer_2021}
\bibfield{author}{\bibinfo{person}{Anna Kramer}.}
  \bibinfo{year}{2021}\natexlab{}.
\newblock \bibinfo{title}{How twitter hired Tech's biggest critics to build
  ethical AI}.
\newblock
\newblock
\urldef\tempurl%
\url{https://www.protocol.com/workplace/twitter-ethical-ai-meta}
\showURL{%
\tempurl}


\bibitem[Kull et~al\mbox{.}(2017)]%
        {Kull2017BetaCA}
\bibfield{author}{\bibinfo{person}{Meelis Kull}, \bibinfo{person}{Telmo
  de~Menezes~e Silva~Filho}, {and} \bibinfo{person}{Peter~A. Flach}.}
  \bibinfo{year}{2017}\natexlab{}.
\newblock \showarticletitle{Beta calibration: a well-founded and easily
  implemented improvement on logistic calibration for binary classifiers}. In
  \bibinfo{booktitle}{\emph{International Conference on Artificial Intelligence
  and Statistics}}.
\newblock


\bibitem[Kull et~al\mbox{.}(2019)]%
        {kull2019betaScaling}
\bibfield{author}{\bibinfo{person}{Meelis Kull}, \bibinfo{person}{Miquel
  Perello~Nieto}, \bibinfo{person}{Markus K{\"a}ngsepp}, \bibinfo{person}{Telmo
  Silva~Filho}, \bibinfo{person}{Hao Song}, {and} \bibinfo{person}{Peter
  Flach}.} \bibinfo{year}{2019}\natexlab{}.
\newblock \showarticletitle{Beyond temperature scaling: Obtaining
  well-calibrated multi-class probabilities with dirichlet calibration}.
\newblock \bibinfo{journal}{\emph{Advances in neural information processing
  systems}}  \bibinfo{volume}{32} (\bibinfo{year}{2019}).
\newblock


\bibitem[Kumar et~al\mbox{.}(2019)]%
        {Kumar2019}
\bibfield{author}{\bibinfo{person}{Ananya Kumar}, \bibinfo{person}{Percy~S.
  Liang}, {and} \bibinfo{person}{Tengyu Ma}.} \bibinfo{year}{2019}\natexlab{}.
\newblock \showarticletitle{Verified Uncertainty Calibration}. In
  \bibinfo{booktitle}{\emph{Advances in Neural Information Processing Systems
  32: Annual Conference on Neural Information Processing Systems 2019, NeurIPS
  2019}}.
\newblock


\bibitem[Kumar et~al\mbox{.}(2018)]%
        {kumar2018trainable}
\bibfield{author}{\bibinfo{person}{Aviral Kumar}, \bibinfo{person}{Sunita
  Sarawagi}, {and} \bibinfo{person}{Ujjwal Jain}.}
  \bibinfo{year}{2018}\natexlab{}.
\newblock \showarticletitle{Trainable calibration measures for neural networks
  from kernel mean embeddings}. In \bibinfo{booktitle}{\emph{International
  Conference on Machine Learning}}. PMLR, \bibinfo{pages}{2805--2814}.
\newblock


\bibitem[Küppers et~al\mbox{.}(2020)]%
        {Kueppers_2020_CVPR_Workshops}
\bibfield{author}{\bibinfo{person}{Fabian Küppers}, \bibinfo{person}{Jan
  Kronenberger}, \bibinfo{person}{Amirhossein Shantia}, {and}
  \bibinfo{person}{Anselm Haselhoff}.} \bibinfo{year}{2020}\natexlab{}.
\newblock \showarticletitle{Multivariate Confidence Calibration for Object
  Detection}. In \bibinfo{booktitle}{\emph{The IEEE/CVF Conference on Computer
  Vision and Pattern Recognition (CVPR) Workshops}}.
\newblock


\bibitem[Lee et~al\mbox{.}(2021)]%
        {LeeSufficiency}
\bibfield{author}{\bibinfo{person}{Joshua~K Lee}, \bibinfo{person}{Yuheng Bu},
  \bibinfo{person}{Deepta Rajan}, \bibinfo{person}{Prasanna Sattigeri},
  \bibinfo{person}{Rameswar Panda}, \bibinfo{person}{Subhro Das}, {and}
  \bibinfo{person}{Gregory~W Wornell}.} \bibinfo{year}{2021}\natexlab{}.
\newblock \showarticletitle{Fair Selective Classification Via Sufficiency}. In
  \bibinfo{booktitle}{\emph{Proceedings of the 38th International Conference on
  Machine Learning}} \emph{(\bibinfo{series}{Proceedings of Machine Learning
  Research}, Vol.~\bibinfo{volume}{139})},
  \bibfield{editor}{\bibinfo{person}{Marina Meila} {and} \bibinfo{person}{Tong
  Zhang}} (Eds.). \bibinfo{publisher}{PMLR}, \bibinfo{pages}{6076--6086}.
\newblock
\urldef\tempurl%
\url{https://proceedings.mlr.press/v139/lee21b.html}
\showURL{%
\tempurl}


\bibitem[Li et~al\mbox{.}(2022)]%
        {li2022more}
\bibfield{author}{\bibinfo{person}{Yunyi Li}, \bibinfo{person}{Maria
  De-Arteaga}, {and} \bibinfo{person}{Maytal Saar-Tsechansky}.}
  \bibinfo{year}{2022}\natexlab{}.
\newblock \showarticletitle{When More Data Lead Us Astray: Active Data
  Acquisition in the Presence of Label Bias}. In
  \bibinfo{booktitle}{\emph{Proceedings of the AAAI Conference on Human
  Computation and Crowdsourcing}}, Vol.~\bibinfo{volume}{10}.
  \bibinfo{pages}{133--146}.
\newblock


\bibitem[Liaw et~al\mbox{.}(2018)]%
        {liaw2018tune}
\bibfield{author}{\bibinfo{person}{Richard Liaw}, \bibinfo{person}{Eric Liang},
  \bibinfo{person}{Robert Nishihara}, \bibinfo{person}{Philipp Moritz},
  \bibinfo{person}{Joseph~E Gonzalez}, {and} \bibinfo{person}{Ion Stoica}.}
  \bibinfo{year}{2018}\natexlab{}.
\newblock \showarticletitle{Tune: A Research Platform for Distributed Model
  Selection and Training}.
\newblock \bibinfo{journal}{\emph{arXiv preprint arXiv:1807.05118}}
  (\bibinfo{year}{2018}).
\newblock


\bibitem[Liu et~al\mbox{.}(2019)]%
        {LiuImplicitFairnessCriterion}
\bibfield{author}{\bibinfo{person}{Lydia~T. Liu}, \bibinfo{person}{Max
  Simchowitz}, {and} \bibinfo{person}{Moritz Hardt}.}
  \bibinfo{year}{2019}\natexlab{}.
\newblock \showarticletitle{The Implicit Fairness Criterion of Unconstrained
  Learning}. In \bibinfo{booktitle}{\emph{Proceedings of the 36th International
  Conference on Machine Learning}} \emph{(\bibinfo{series}{Proceedings of
  Machine Learning Research}, Vol.~\bibinfo{volume}{97})},
  \bibfield{editor}{\bibinfo{person}{Kamalika Chaudhuri} {and}
  \bibinfo{person}{Ruslan Salakhutdinov}} (Eds.). \bibinfo{publisher}{PMLR},
  \bibinfo{pages}{4051--4060}.
\newblock
\urldef\tempurl%
\url{https://proceedings.mlr.press/v97/liu19f.html}
\showURL{%
\tempurl}


\bibitem[Logan(2022)]%
        {logan_2022}
\bibfield{author}{\bibinfo{person}{Heloise Logan}.}
  \bibinfo{year}{2022}\natexlab{}.
\newblock \bibinfo{title}{A closer look at how linkedin integrates fairness
  into its AI products}.
\newblock
\newblock
\urldef\tempurl%
\url{https://engineering.linkedin.com/blog/2022/a-closer-look-at-how-linkedin-integrates-fairness-into-its-ai-pr}
\showURL{%
\tempurl}


\bibitem[Loi and Heitz(2022)]%
        {Loi2022IsCA}
\bibfield{author}{\bibinfo{person}{Michele Loi} {and}
  \bibinfo{person}{Christoph Heitz}.} \bibinfo{year}{2022}\natexlab{}.
\newblock \showarticletitle{Is calibration a fairness requirement?: An argument
  from the point of view of moral philosophy and decision theory}.
\newblock \bibinfo{journal}{\emph{2022 ACM Conference on Fairness,
  Accountability, and Transparency}} (\bibinfo{year}{2022}).
\newblock


\bibitem[Mendez et~al\mbox{.}(2022)]%
        {Mendez2022-MENEIO}
\bibfield{author}{\bibinfo{person}{Julian~Alfredo Mendez},
  \bibinfo{person}{R\"{u}ya~G\"{o}khan Ko\c{c}er}, \bibinfo{person}{Flavia
  Barsotti}, {and} \bibinfo{person}{Andrea~Aler Tubella}.}
  \bibinfo{year}{2022}\natexlab{}.
\newblock \showarticletitle{Ethical Implications of Fairness Interventions:
  What Might Be Hidden Behind Engineering Choices?}
\newblock \bibinfo{journal}{\emph{Ethics and Information Technology}}
  \bibinfo{volume}{24}, \bibinfo{number}{1} (\bibinfo{year}{2022}).
\newblock
\urldef\tempurl%
\url{https://doi.org/10.1007/s10676-022-09636-z}
\showDOI{\tempurl}


\bibitem[Minderer et~al\mbox{.}(2021)]%
        {minderer2021revisiting}
\bibfield{author}{\bibinfo{person}{Matthias Minderer}, \bibinfo{person}{Josip
  Djolonga}, \bibinfo{person}{Rob Romijnders}, \bibinfo{person}{Frances Hubis},
  \bibinfo{person}{Xiaohua Zhai}, \bibinfo{person}{Neil Houlsby},
  \bibinfo{person}{Dustin Tran}, {and} \bibinfo{person}{Mario Lucic}.}
  \bibinfo{year}{2021}\natexlab{}.
\newblock \showarticletitle{Revisiting the calibration of modern neural
  networks}.
\newblock \bibinfo{journal}{\emph{Advances in Neural Information Processing
  Systems}}  \bibinfo{volume}{34} (\bibinfo{year}{2021}),
  \bibinfo{pages}{15682--15694}.
\newblock


\bibitem[Mukhoti et~al\mbox{.}(2020)]%
        {mukhoti2020calibrating}
\bibfield{author}{\bibinfo{person}{Jishnu Mukhoti}, \bibinfo{person}{Viveka
  Kulharia}, \bibinfo{person}{Amartya Sanyal}, \bibinfo{person}{Stuart
  Golodetz}, \bibinfo{person}{Philip Torr}, {and} \bibinfo{person}{Puneet
  Dokania}.} \bibinfo{year}{2020}\natexlab{}.
\newblock \showarticletitle{Calibrating deep neural networks using focal loss}.
\newblock \bibinfo{journal}{\emph{Advances in Neural Information Processing
  Systems}}  \bibinfo{volume}{33} (\bibinfo{year}{2020}),
  \bibinfo{pages}{15288--15299}.
\newblock


\bibitem[Murphy(1986)]%
        {murphy1986new}
\bibfield{author}{\bibinfo{person}{Allan~H Murphy}.}
  \bibinfo{year}{1986}\natexlab{}.
\newblock \showarticletitle{A new decomposition of the Brier score: Formulation
  and interpretation}.
\newblock \bibinfo{journal}{\emph{Monthly weather review}}
  \bibinfo{volume}{114}, \bibinfo{number}{12} (\bibinfo{year}{1986}),
  \bibinfo{pages}{2671--2673}.
\newblock


\bibitem[Naeini et~al\mbox{.}(2015)]%
        {naeini2015bbq}
\bibfield{author}{\bibinfo{person}{Mahdi~Pakdaman Naeini},
  \bibinfo{person}{Gregory Cooper}, {and} \bibinfo{person}{Milos Hauskrecht}.}
  \bibinfo{year}{2015}\natexlab{}.
\newblock \showarticletitle{Obtaining well calibrated probabilities using
  bayesian binning}. In \bibinfo{booktitle}{\emph{Twenty-Ninth AAAI Conference
  on Artificial Intelligence}}.
\newblock


\bibitem[Naeini and Cooper(2016)]%
        {naeini2016ENIR}
\bibfield{author}{\bibinfo{person}{Mahdi~Pakdaman Naeini} {and}
  \bibinfo{person}{Gregory~F Cooper}.} \bibinfo{year}{2016}\natexlab{}.
\newblock \showarticletitle{Binary classifier calibration using an ensemble of
  near isotonic regression models}. In \bibinfo{booktitle}{\emph{2016 IEEE 16th
  International Conference on Data Mining (ICDM)}}. IEEE,
  \bibinfo{pages}{360--369}.
\newblock


\bibitem[Nandy(2021)]%
        {nandy_2021}
\bibfield{author}{\bibinfo{person}{Preetam Nandy}.}
  \bibinfo{year}{2021}\natexlab{}.
\newblock \bibinfo{title}{Using the linkedin fairness toolkit in large-scale AI
  systems}.
\newblock
\newblock
\urldef\tempurl%
\url{https://engineering.linkedin.com/blog/2021/using-the-linkedin-fairness-toolkit-large-scale-ai}
\showURL{%
\tempurl}


\bibitem[Nandy et~al\mbox{.}(2020)]%
        {NandyEOdds}
\bibfield{author}{\bibinfo{person}{Preetam Nandy}, \bibinfo{person}{Cyrus
  Diciccio}, \bibinfo{person}{Divya Venugopalan}, \bibinfo{person}{Heloise
  Logan}, \bibinfo{person}{Kinjal Basu}, {and} \bibinfo{person}{Noureddine~El
  Karoui}.} \bibinfo{year}{2020}\natexlab{}.
\newblock \bibinfo{title}{Achieving Fairness via Post-Processing in Web-Scale
  Recommender Systems}.
\newblock
\newblock
\urldef\tempurl%
\url{https://doi.org/10.48550/ARXIV.2006.11350}
\showDOI{\tempurl}


\bibitem[Niculescu-Mizil and Caruana(2005)]%
        {MizilPlattIsotonic}
\bibfield{author}{\bibinfo{person}{Alexandru Niculescu-Mizil} {and}
  \bibinfo{person}{Rich Caruana}.} \bibinfo{year}{2005}\natexlab{}.
\newblock \showarticletitle{Predicting Good Probabilities with Supervised
  Learning}. In \bibinfo{booktitle}{\emph{Proceedings of the 22nd International
  Conference on Machine Learning}} (Bonn, Germany) \emph{(\bibinfo{series}{ICML
  '05})}. \bibinfo{publisher}{Association for Computing Machinery},
  \bibinfo{address}{New York, NY, USA}, \bibinfo{pages}{625–632}.
\newblock
\showISBNx{1595931805}
\urldef\tempurl%
\url{https://doi.org/10.1145/1102351.1102430}
\showDOI{\tempurl}


\bibitem[Paszke et~al\mbox{.}(2019)]%
        {pytorch}
\bibfield{author}{\bibinfo{person}{Adam Paszke}, \bibinfo{person}{Sam Gross},
  \bibinfo{person}{Francisco Massa}, \bibinfo{person}{Adam Lerer},
  \bibinfo{person}{James Bradbury}, \bibinfo{person}{Gregory Chanan},
  \bibinfo{person}{Trevor Killeen}, \bibinfo{person}{Zeming Lin},
  \bibinfo{person}{Natalia Gimelshein}, \bibinfo{person}{Luca Antiga},
  \bibinfo{person}{Alban Desmaison}, \bibinfo{person}{Andreas Köpf},
  \bibinfo{person}{Edward Yang}, \bibinfo{person}{Zach DeVito},
  \bibinfo{person}{Martin Raison}, \bibinfo{person}{Alykhan Tejani},
  \bibinfo{person}{Sasank Chilamkurthy}, \bibinfo{person}{Benoit Steiner},
  \bibinfo{person}{Lu Fang}, \bibinfo{person}{Junjie Bai}, {and}
  \bibinfo{person}{Soumith Chintala}.} \bibinfo{year}{2019}\natexlab{}.
\newblock \bibinfo{title}{PyTorch: An Imperative Style, High-Performance Deep
  Learning Library}.
\newblock
\newblock
\urldef\tempurl%
\url{https://doi.org/10.48550/ARXIV.1912.01703}
\showDOI{\tempurl}


\bibitem[Pfohl et~al\mbox{.}(2021)]%
        {pfohl2021empirical}
\bibfield{author}{\bibinfo{person}{Stephen~R Pfohl}, \bibinfo{person}{Agata
  Foryciarz}, {and} \bibinfo{person}{Nigam~H Shah}.}
  \bibinfo{year}{2021}\natexlab{}.
\newblock \showarticletitle{An empirical characterization of fair machine
  learning for clinical risk prediction}.
\newblock \bibinfo{journal}{\emph{Journal of biomedical informatics}}
  \bibinfo{volume}{113} (\bibinfo{year}{2021}), \bibinfo{pages}{103621}.
\newblock


\bibitem[Platt and Karampatziakis(2007)]%
        {Platt2007ProbabilisticOF}
\bibfield{author}{\bibinfo{person}{John Platt} {and} \bibinfo{person}{Nikos
  Karampatziakis}.} \bibinfo{year}{2007}\natexlab{}.
\newblock \showarticletitle{Probabilistic Outputs for SVMs and Comparisons to
  Regularized Likelihood Methods}.
\newblock


\bibitem[Platt(1999)]%
        {Platt99probabilisticoutputs}
\bibfield{author}{\bibinfo{person}{John~C. Platt}.}
  \bibinfo{year}{1999}\natexlab{}.
\newblock \showarticletitle{Probabilistic Outputs for Support Vector Machines
  and Comparisons to Regularized Likelihood Methods}. In
  \bibinfo{booktitle}{\emph{ADVANCES IN LARGE MARGIN CLASSIFIERS}}.
  \bibinfo{publisher}{MIT Press}, \bibinfo{pages}{61--74}.
\newblock


\bibitem[Pleiss et~al\mbox{.}(2017)]%
        {pleiss2017fairness}
\bibfield{author}{\bibinfo{person}{Geoff Pleiss}, \bibinfo{person}{Manish
  Raghavan}, \bibinfo{person}{Felix Wu}, \bibinfo{person}{Jon Kleinberg}, {and}
  \bibinfo{person}{Kilian~Q Weinberger}.} \bibinfo{year}{2017}\natexlab{}.
\newblock \showarticletitle{On fairness and calibration}. In
  \bibinfo{booktitle}{\emph{Advances in Neural Information Processing
  Systems}}. \bibinfo{pages}{5680--5689}.
\newblock


\bibitem[Rambachan et~al\mbox{.}(2020)]%
        {rambachan2020economic}
\bibfield{author}{\bibinfo{person}{Ashesh Rambachan}, \bibinfo{person}{Jon
  Kleinberg}, \bibinfo{person}{Jens Ludwig}, {and} \bibinfo{person}{Sendhil
  Mullainathan}.} \bibinfo{year}{2020}\natexlab{}.
\newblock \showarticletitle{An economic perspective on algorithmic fairness}.
  In \bibinfo{booktitle}{\emph{AEA Papers and Proceedings}},
  Vol.~\bibinfo{volume}{110}. \bibinfo{pages}{91--95}.
\newblock


\bibitem[Rezaei et~al\mbox{.}(2020)]%
        {Rezaei_2020}
\bibfield{author}{\bibinfo{person}{Ashkan Rezaei}, \bibinfo{person}{Rizal
  Fathony}, \bibinfo{person}{Omid Memarrast}, {and} \bibinfo{person}{Brian
  Ziebart}.} \bibinfo{year}{2020}\natexlab{}.
\newblock \showarticletitle{Fairness for Robust Log Loss Classification}.
\newblock \bibinfo{journal}{\emph{Proceedings of the {AAAI} Conference on
  Artificial Intelligence}} \bibinfo{volume}{34}, \bibinfo{number}{04}
  (\bibinfo{date}{apr} \bibinfo{year}{2020}), \bibinfo{pages}{5511--5518}.
\newblock
\urldef\tempurl%
\url{https://doi.org/10.1609/aaai.v34i04.6002}
\showDOI{\tempurl}


\bibitem[Rodolfa et~al\mbox{.}(2021)]%
        {Rodolfa_2021}
\bibfield{author}{\bibinfo{person}{Kit~T. Rodolfa}, \bibinfo{person}{Hemank
  Lamba}, {and} \bibinfo{person}{Rayid Ghani}.}
  \bibinfo{year}{2021}\natexlab{}.
\newblock \showarticletitle{Empirical observation of negligible
  fairness{\textendash}accuracy trade-offs in machine learning for public
  policy}.
\newblock \bibinfo{journal}{\emph{Nature Machine Intelligence}}
  \bibinfo{volume}{3}, \bibinfo{number}{10} (\bibinfo{date}{oct}
  \bibinfo{year}{2021}), \bibinfo{pages}{896--904}.
\newblock
\urldef\tempurl%
\url{https://doi.org/10.1038/s42256-021-00396-x}
\showDOI{\tempurl}


\bibitem[Roelofs et~al\mbox{.}(2022)]%
        {roelofs2022mitigating}
\bibfield{author}{\bibinfo{person}{Rebecca Roelofs}, \bibinfo{person}{Nicholas
  Cain}, \bibinfo{person}{Jonathon Shlens}, {and} \bibinfo{person}{Michael~C
  Mozer}.} \bibinfo{year}{2022}\natexlab{}.
\newblock \showarticletitle{Mitigating bias in calibration error estimation}.
  In \bibinfo{booktitle}{\emph{International Conference on Artificial
  Intelligence and Statistics}}. PMLR, \bibinfo{pages}{4036--4054}.
\newblock


\bibitem[Sagawa et~al\mbox{.}(2019)]%
        {sagawa2019distributionally}
\bibfield{author}{\bibinfo{person}{Shiori Sagawa}, \bibinfo{person}{Pang~Wei
  Koh}, \bibinfo{person}{Tatsunori~B Hashimoto}, {and} \bibinfo{person}{Percy
  Liang}.} \bibinfo{year}{2019}\natexlab{}.
\newblock \showarticletitle{Distributionally Robust Neural Networks}. In
  \bibinfo{booktitle}{\emph{International Conference on Learning
  Representations}}.
\newblock


\bibitem[Xiang(2020)]%
        {xiang2020reconciling}
\bibfield{author}{\bibinfo{person}{Alice Xiang}.}
  \bibinfo{year}{2020}\natexlab{}.
\newblock \showarticletitle{Reconciling legal and technical approaches to
  algorithmic bias}.
\newblock \bibinfo{journal}{\emph{Tenn. L. Rev.}}  \bibinfo{volume}{88}
  (\bibinfo{year}{2020}), \bibinfo{pages}{649}.
\newblock


\bibitem[Zadrozny and Elkan(2002)]%
        {isotonicReg}
\bibfield{author}{\bibinfo{person}{Bianca Zadrozny} {and}
  \bibinfo{person}{Charles Elkan}.} \bibinfo{year}{2002}\natexlab{}.
\newblock \showarticletitle{Transforming Classifier Scores into Accurate
  Multiclass Probability Estimates} \emph{(\bibinfo{series}{KDD '02})}.
  \bibinfo{publisher}{Association for Computing Machinery},
  \bibinfo{address}{New York, NY, USA}, \bibinfo{pages}{694–699}.
\newblock
\showISBNx{158113567X}
\urldef\tempurl%
\url{https://doi.org/10.1145/775047.775151}
\showDOI{\tempurl}


\bibitem[Zadrozny and Elkan(2001)]%
        {Zadrozny2001ObtainingCP}
\bibfield{author}{\bibinfo{person}{Bianca Zadrozny} {and}
  \bibinfo{person}{Charles~Peter Elkan}.} \bibinfo{year}{2001}\natexlab{}.
\newblock \showarticletitle{Obtaining calibrated probability estimates from
  decision trees and naive Bayesian classifiers}. In
  \bibinfo{booktitle}{\emph{International Conference on Machine Learning}}.
\newblock


\bibitem[Zeng et~al\mbox{.}(2022)]%
        {zeng2022sufficiency}
\bibfield{author}{\bibinfo{person}{Xianli Zeng}, \bibinfo{person}{Edgar
  Dobriban}, {and} \bibinfo{person}{Guang Cheng}.}
  \bibinfo{year}{2022}\natexlab{}.
\newblock \showarticletitle{Fair Bayes-Optimal Classifiers Under Predictive
  Parity}.
\newblock \bibinfo{journal}{\emph{arXiv preprint arXiv:2205.07182}}
  (\bibinfo{year}{2022}).
\newblock


\bibitem[Zhai et~al\mbox{.}(2021)]%
        {zhai2021doro}
\bibfield{author}{\bibinfo{person}{Runtian Zhai}, \bibinfo{person}{Chen Dan},
  \bibinfo{person}{Zico Kolter}, {and} \bibinfo{person}{Pradeep Ravikumar}.}
  \bibinfo{year}{2021}\natexlab{}.
\newblock \showarticletitle{Doro: Distributional and outlier robust
  optimization}. In \bibinfo{booktitle}{\emph{International Conference on
  Machine Learning}}. PMLR, \bibinfo{pages}{12345--12355}.
\newblock


\bibitem[{\v{Z}}liobait{\.e} and Custers(2016)]%
        {vzliobaite2016using}
\bibfield{author}{\bibinfo{person}{Indr{\.e} {\v{Z}}liobait{\.e}} {and}
  \bibinfo{person}{Bart Custers}.} \bibinfo{year}{2016}\natexlab{}.
\newblock \showarticletitle{Using sensitive personal data may be necessary for
  avoiding discrimination in data-driven decision models}.
\newblock \bibinfo{journal}{\emph{Artificial Intelligence and Law}}
  \bibinfo{volume}{24} (\bibinfo{year}{2016}), \bibinfo{pages}{183--201}.
\newblock


\end{thebibliography}
